\pdfoutput=1
\documentclass[11pt]{article}
\usepackage[final]{acl}

\usepackage{times}
\usepackage{latexsym}
\usepackage{float}
\usepackage[T1]{fontenc}

\usepackage[utf8]{inputenc}

\usepackage{microtype}
\usepackage{pdfpages}

\usepackage{inconsolata}
\usepackage{booktabs}
\usepackage{multirow}
\usepackage{svg}
\usepackage{graphicx}

\usepackage{hyperref}

\makeatletter
\def\thanks#1{\protected@xdef\@thanks{\@thanks
        \protect\footnotetext{#1}}}
\makeatother

\title{Unraveling the Truth: Do VLMs really Understand Charts? \\A Deep Dive into Consistency and Robustness}

\author{
Srija Mukhopadhyay\textsuperscript{*}, 
Adnan Qidwai\textsuperscript{*},
Aparna Garimella\textsuperscript{\dag}, 
Pritika Ramu\textsuperscript{\dag}\\
\textbf{Vivek Gupta}\thanks{~*~contributed equally, \ddag~corresponding author (work done at UPenn)}\textsuperscript{\ddag}, 
\textbf{Dan Roth}\textsuperscript{§} \\
\textsuperscript{*}IIIT Hyderabad,
\textsuperscript{\dag}Adobe Research,
\textsuperscript{\ddag}Arizona State University,
\textsuperscript{§}University of Pennsylvania\\
\texttt{\small \{srija.mukhopadhyay@research, adnan.qidwai@students\}.iiit.ac.in}, \\
\texttt{\small \{garimell,pramu\}@adobe.com}; 
\texttt{\small vgupt140@asu.edu};
\texttt{\small danroth@seas.upenn.edu} \\
}

\begin{document}
\maketitle

\begin{abstract}
Chart question answering (CQA) is a crucial area of Visual Language Understanding. However, the robustness and consistency of current Visual Language Models (VLMs) in this field remain under-explored. This paper evaluates state-of-the-art VLMs on comprehensive datasets, developed specifically for this study, encompassing diverse question categories and chart formats. We investigate two key aspects: 1) the models' ability to handle varying levels of chart and question complexity, and 2) their robustness across different visual representations of the same underlying data. Our analysis reveals significant performance variations based on question and chart types, highlighting both strengths and weaknesses of current models. Additionally, we identify areas for improvement and propose future research directions to build more robust and reliable CQA systems. This study sheds light on the limitations of current models and paves the way for future advancements in the field.

\end{abstract}

\section{Introduction} 
Chart question answering (CQA) \cite{masry-etal-2022-chartqa, chaudhry2020leaf} has emerged as a critical area within the field of Visual Language Understanding (VLU) \cite{lee2023pix2struct, ghosh2024exploringfrontiervisionlanguagemodels}, aiming to equip machines with the ability to comprehend and answer questions based on data visualizations. While recent advancements in Vision Language Models (VLMs) and Multimodal Large Language Models (MLLMs) have yielded impressive performance improvements in CQA \cite{liu-etal-2023-matcha, masry-etal-2023-unichart, xia2024chartx, xu2024chartbench, team2023gemini, achiam2023gpt, meng2024chartassisstant}, their true capabilities remain obscure in uncertainty. This paper delves into an insightful analysis of the robustness and consistency of state-of-the-art CQA models, exposing their limitations and guiding future research directions.

\begin{figure}
    \centering
    \includegraphics[width=\linewidth]{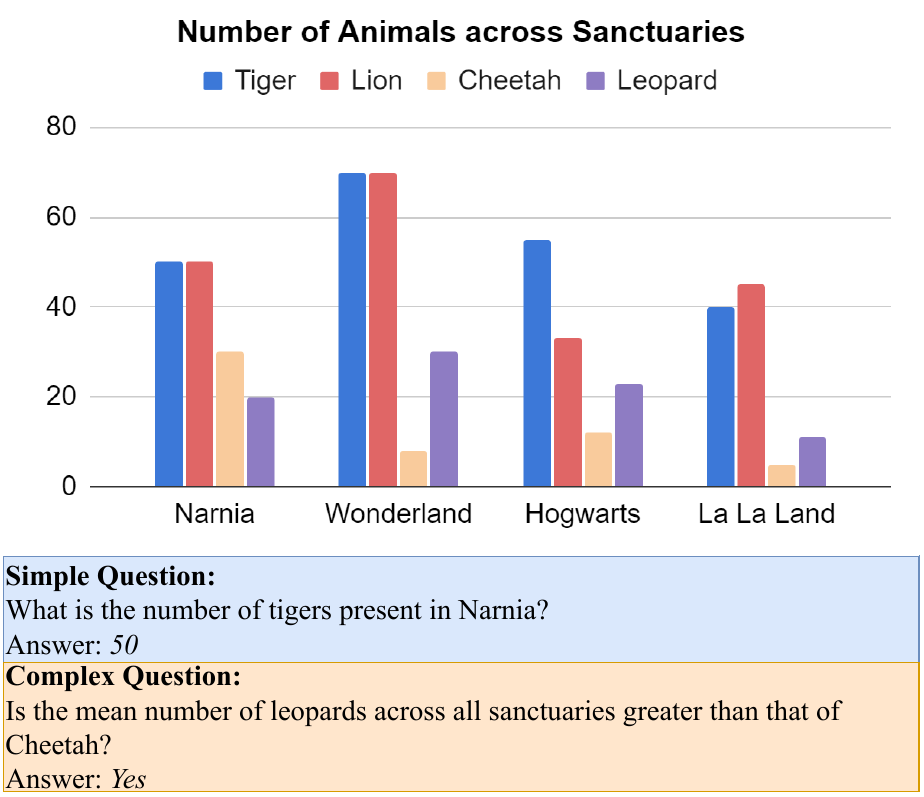}
    \vspace{-1.5em}
    \caption{\small Simple and Complex Questions on a Complex chart}
    \label{fig:example-ques}
    \vspace{-1.5em}
\end{figure}

We address several key questions regarding the current state of CQA: \textbf{Are existing models truly effective, or do their impressive average scores mask significant weaknesses?} For instance, in Figure \ref{fig:example-ques}, one can ask that if the model's performance remains consistent across two distinct question types? The first type, \emph{Simple Questions} like \emph{"What is the number of tigers present in Narnia?"}, involves straightforward value extraction. In contrast, \emph{Complex Questions} such as \emph{"Is the mean number of leopards across all sanctuaries greater than that of cheetah?"} require extracting multiple values, aggregating them, and making boolean comparisons. It's evident that complex questions pose challenges even for humans; understanding how models handle these complexities provides valuable insights into their capabilities.

\begin{figure}
    \centering
    \includegraphics[width = 0.90\linewidth]{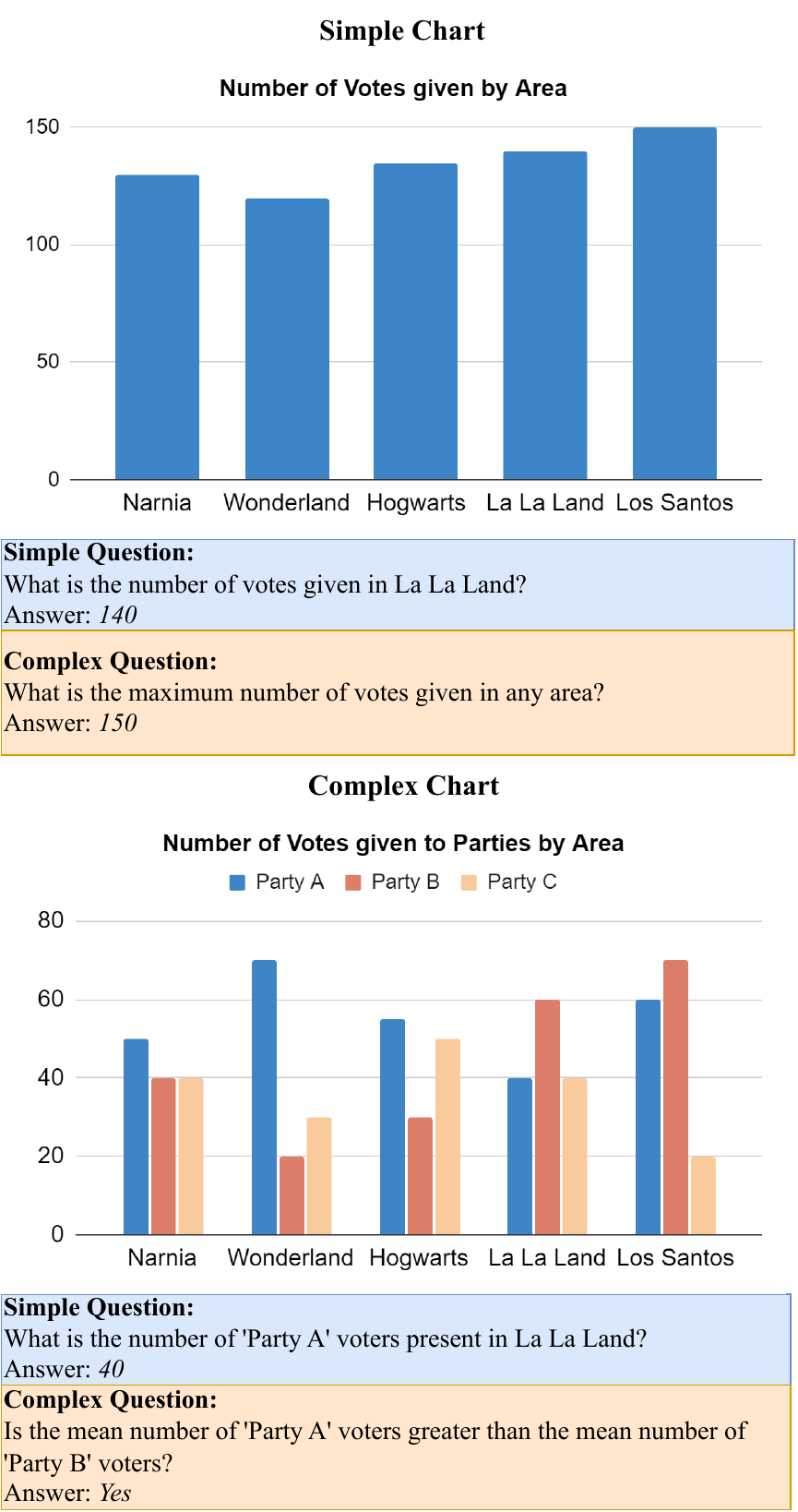}
    \vspace{-0.5em}
    \caption{\small Example of simple chart and complex chart, along with simple and complex questions.}
    \label{fig:div-examples}
    \vspace{-1.5em}
\end{figure}

\textbf{How do models perform on specific aspects of chart understanding, such as question complexity and chart type?} Figure \ref{fig:div-examples} shows the different types of charts across which the performance of a model can be evaluated—specifically, \emph{Simple Charts} and \emph{Complex Charts}—along with the different possible question types, including \emph{Simple Questions} and \emph{Complex Questions}. Complex Charts, such as grouped-bar charts that compare multiple attributes side by side present information in a more intricate manner compared to Simple Charts, which depict data about a single attribute using a single bar. Similarly, questions can range from complex tasks like identifying maximum values, performing aggregations, and making comparisons, to simpler queries focused on straightforward value extraction. Investigating how models handle these varied chart types and question complexities provides crucial insights into their performance and adaptability.

\textbf{Furthermore, is the robustness of these models, their ability to generalize across diverse variations, adequately explored?} The same data can be depicted in multiple visual formats. For instance, Figure \ref{fig:perturbations} demonstrates how an original chart can be transformed into stair plots, bar charts, stacked representations, and many more. These variations can differ in aspects such as color schemes, patterns, legend positioning, and even details specific to each chart type like legend orientation and grid sizes on the x-axis and y-axis. Exploring the effect of these variations could provide deeper insights into the data and enhance the comprehensibility of the visualizations for models.

To answer these questions, we present a rigorous evaluation of leading CQA models on a meticulously curated dataset. This dataset includes diverse chart types and question categories, allowing for a thorough assessment of model performance across varying levels of complexity. We examine how well the models generalize across diverse visual representations of identical data, assessing their robustness against perturbations.

Our findings reveal significant performance discrepancies, particularly when transitioning from simple to complex chart-question combinations. Moreover, we demonstrate that even the highest-performing models exhibit a substantial drop in accuracy when subjected to diverse perturbations, highlighting the critical need for improved robustness in CQA.
This paper makes the following contributions:
\begin{itemize}
\setlength\itemsep{0.0em}
    \item Providing a thorough analysis of the strengths and weaknesses of current VLMs and MLLMs for chart understanding.
    \item Introducing a new evaluation set with fine-grained splits across chart types and question complexities, facilitating a deeper understanding of model performance.
    \item Performing a detailed robustness analysis to uncover the shortcomings of current models, emphasizing the necessity for additional research in this domain.
\end{itemize}
Our research sheds light on the current state of CQA, offering crucial insights. Our datasets, along with all the associated scripts, are available at \href{https://robustcqa.github.io/}{https://robustcqa.github.io/}

\section{Initial Dataset} 
This section highlights the dataset preparation process employed to analyze the performance of CQA models across a spectrum of chart types and question complexities. 

\subsection{Dataset Selection}
To ensure a comprehensive evaluation of CQA models, we selected the ChartQA dataset \cite{masry-etal-2022-chartqa} as our primary benchmark. This dataset is widely used in CQA benchmarking, covering diverse domains from sources like Our World in Data, Statista, OECD, and Pew Research.

ChartQA includes two distinct question categories: \emph{"Human"} and \emph{"Augmented"}. \emph{"Human"} questions were generated by human annotators, while \emph{"Augmented"} questions were machine-generated, ensuring a diverse spectrum of question styles. Another important aspect which motivated our choice of ChartQA dataset was the presence of underlying tables. This feature enabled us to generate controlled visual perturbations for the later section of our study.  Our experiments were conducted exclusively on the test set of ChartQA, comprising questions, charts and the corresponding tables.

\subsection{Chart and Question Labelling}
To facilitate a more granular analysis of model performance, we categorized both charts and questions according to their complexity levels. This categorization was applied to the entire ChartQA test set, resulting in a modified evaluation dataset tailored for our experiments.

\paragraph{Chart Categorization.} The tables provided by ChartQA were loaded as a pandas dataframe. We classify charts as either simple or complex, based on column count in the dataframe: two columns indicate a simple chart, while more than two columns signify a more complex chart. We leverage this fact to classify them using a python script.
    
- \textbf{Simple Charts:} The tables of these charts contain two columns to represent the dependent and independent variable. The charts thus formed represent a single entity and exhibit no overlaps or complex visual elements. Figure \ref{fig:div-examples} shows an example of such chart titled \emph{"Number of Votes given by Area".}
    
- \textbf{Complex Charts:} The tables of these charts feature more than two columns, often encompassing multiple dependent variables. Thus, the charts formed have increased visual complexity. These charts usually depict multiple entities over a common series. Figure \ref{fig:div-examples} shows an example of such chart titled \emph{"Number of Votes given to Parties by Area".}

\paragraph{Question Categorization.} Human annotators cleaned and categorized the questions from the ChartQA dataset into two categories based on their complexity: 
    
    - \textbf{Simple Questions:} These questions primarily focus on data extraction, and typically involve a \textbf{single step of reasoning}. A human annotator can ideally answer such a question in a single step. Figure \ref{fig:div-examples} shows an example of such questions \emph{"What is the number of votes given in La La La Land?"}. One can simply answer the question by fetching the value from the chart.
    
    - \textbf{Complex Questions:} These questions require multi-step reasoning along with data extraction, and often involve comparisons and logical inferences. If it takes multiple steps for the human annotator to answer a question, it would be classified as a complex question. Figure \ref{fig:div-examples} shows an example of such questions \emph{"Is the mean number of `Party A' voters greater than the mean number of `Party B' voters?"}. For this question, one would require multiple calculations to reach the final answer.

We introduced these categorizations while preserving the existing division of question generation types (human-generated and augmented questions) which was present in the original dataset, resulting in eight categories. The number of unique question-chart pairs in each category is presented in Table \ref{tab:dataset-div}. We call this modified dataset ChartQA-Split. The detailed catagorization in our dataset allows us to isolate the impact of chart and question complexity on model performance, providing a deeper understanding of their capabilities and limitations. 

\begin{table}[H]
\vspace{-0.5em}
\centering
\small
\begin{tabular}{@{}ccc|cc@{}}
\toprule
        & \multicolumn{2}{c|}{Human} & \multicolumn{2}{c}{Augmented} \\ \midrule
        & Simple      & Complex      & Simple        & Complex       \\ \midrule
Simple  & 149         & 450          & 876           & 165           \\
Complex & 143         & 419          & 133           & 38            \\ \bottomrule
\end{tabular}%
\vspace{-0.75em}
\caption{Dataset statistics. Rows represent the type of Chart, Columns represent the type of Question and its Generation method. }
\label{tab:dataset-div}
\vspace{-0.75em}
\end{table}

\section{Experiments}
\paragraph{Models.} To rigorously assess the performance of CQA models, we selected a diverse range of state-of-the-art models, varying in architecture, size, and training setup. All models were evaluated using a zero-shot Chain-of-Thought \cite{wei2022cot} prompting approach. An example of our prompt can be found in Figure \ref{fig:example_prompt_qa}. It is important to note that no additional reasoning aids were provided to any of the models. For the sake of clarity and analysis, we grouped the models into three broad categories: 

\paragraph{Chart-based VLMs.} This category contains open-source VLMs specifically adapted for chart reasoning. 
\emph{MatCha (282M)} \cite{liu-etal-2023-matcha} is a transformer based model which enhances the capabilities of Pix2Struct \cite{lee2023pix2struct} models through pre-training on mathematical reasoning and chart derendering tasks.
\emph{UniChart (201M)} \cite{masry-etal-2023-unichart} is another similar model which achieves chart understanding by leveraging pre-training on tasks such as data table generation, numerical and visual reasoning, and open-ended question answering. 
\emph{DePlot (282M)} \cite{liu-etal-2023-deplot} is a model which specializes on extracting tabular data from a given chart. The extracted table is subsequently passed to a Language Model (LM), e.g. \emph{Flan UL2 (20B)} \cite{Tay2022UL2UL}, for reasoning via Chain-of-Thought prompting \cite{wei2022cot}. 

\paragraph{Generalist VLMs.}This category comprises open-source VLMs trained on general visual comprehension tasks. Notably, these models were not specifically trained or adapted for chart reasoning.
\emph{QwenVL} \cite{Qwen-VL} is a  generalist 7-billion-parameter VLM built on top of \emph{Qwen-LM} \cite{qwen} through the integration of visual encoders and the use of general and multi-task pre-training.
\emph{CogAgent VQA} \cite{hong2024cogagent} is an 18-billion-parameter VLM specializing in Graphical User Interface (GUI) understanding and navigation.
\emph{InternLM-XComposer2 (8B)} \cite{internlmxcomposer2} is an adaptation of \emph{InternLM2-7B} \cite{cai2024internlm2}, excelling in producing high-quality long-text multi-modal content and reasoning within visual-language understanding contexts. 

\paragraph{Large MLLMs.} This category features state-of-the-art closed-source Multimodal Large Language Models (MLLMs) pre-trained on extensive visual and language data.  
For this category, we utilized \emph{Gemini 1.5 Flash} \cite{team2023gemini}, and \emph{GPT-4o} \cite{achiam2023gpt}, renowned for their capabilities in reasoning and visual understanding. 

\paragraph{Evaluation} To evaluate our models, we decided to utilize the Relaxed Accuracy metric owing to the objective nature of the expected answers. To improve on the Relaxed Accuracy metric, we introduce extra checks for precise and accurate answer matching. This metric, similar to Relaxed Accuracy, provides a 5\% leverage for numerical answer 
matching. However, it includes the following checks:

    \textbf{Alphanumeric String Matching:} Removing comma and spaces from the given answer and gold label to ensure an exact alphanumeric string comparison. 
    
    \textbf{Strict Year Matching:} For questions specifically asking for a "Year" as an answer, the 5\% relaxation is disabled, forcing a strict string match. This ensures that the model accurately identifies the correct year. 
    
    \textbf{Unordered Exact List Matching:} For questions requiring multiple answers, an unordered exact list matching is applied, to ensure that the model correctly identifies all the expected elements in answer list, regardless of their order.

Furthermore, to validate the accuracy of our proposed evaluation metric, we manually verified the answers obtained using this metric. Our metric is usable and applicable for general large-scale model evaluation in question-answering based tasks. 

\paragraph{Smaller VLMs.} We noticed that smaller models (QwenVL, CogAgent, InternLM) struggled to produce answers in the correct format. This might be due lack of complex instruction following abilities. We addressed this by using Gemini 1.5 Flash to extract answers from their outputs in a favourable format, hence using the \textit{LLM as an extractor}. \textbf{Manual verification} of 150 samples confirmed that Gemini 1.5 Flash primarily acted as a formatting tool, preserving the original model's answer in 149 cases and performing rounding in the one remaining instance. This demonstrates Gemini's effectiveness in enhancing the usability of smaller models without significantly altering their intent. The prompt used, can be found in Figure \ref{fig:extractorPrompt}.

\section{Can models reasons consistently?}
This section presents our findings and analysis on the performance of various chart question answering (CQA) models across different chart types and question complexities. 

\subsection{Results and Discussion}\label{results}
Table \ref{tab:initial-score} gives an overview of all results obtained for this section.

\begin{table}[!htb]
\scriptsize
\setlength{\tabcolsep}{2.7pt}
\centering
\begin{tabular}{@{}ccccccccc@{}}
\toprule
Type & \multicolumn{3}{c|}{Chart-based VLMs}                                                                 & \multicolumn{3}{c|}{Generalist VLMs}                                                                                                                  & \multicolumn{2}{c}{MLLMs}                                                    \\ \midrule
       & \begin{tabular}[c]{@{}c@{}}Mat-\\Cha\end{tabular} & \begin{tabular}[c]{@{}c@{}}Uni-\\Chart\end{tabular} & \multicolumn{1}{c|}{\begin{tabular}[c]{@{}c@{}}DePlot + \\ Flan UL2\end{tabular}} & \begin{tabular}[c]{@{}c@{}}Qwen\\VL\end{tabular} & \begin{tabular}[c]{@{}c@{}}CogAgent\\ VQA\end{tabular} & \multicolumn{1}{c|}{\begin{tabular}[c]{@{}c@{}}Intern\\LM\end{tabular}} & \begin{tabular}[c]{@{}c@{}}Gemini \\ 1.5 Flash\end{tabular} & \begin{tabular}[c]{@{}c@{}}GPT \\ 4o \end{tabular}        \\ \midrule
\multicolumn{9}{c}{Human}                                                                                                                                                                                                                                                                                                                             \\ \midrule
SS     & 57.00  & 49.60    & \multicolumn{1}{c|}{51.60}                                                        & 66.40   & 81.20                                                  & \multicolumn{1}{c|}{79.90}                                                         & 87.92                                                       & \textbf{88.59} \\
SC     & 30.22  & 32.00    & \multicolumn{1}{c|}{32.80}                                                        & 44.20   & 55.50                                                  & \multicolumn{1}{c|}{58.60}                                                         & 81.11                                                       & \textbf{88.22} \\
CS     & 45.40  & 47.50    & \multicolumn{1}{c|}{30.60}                                                        & 60.10   & 58.00                                                  & \multicolumn{1}{c|}{74.10}                                                         & 80.42                                                       & \textbf{81.82} \\
CC     & 25.29  & 25.00    & \multicolumn{1}{c|}{25.20}                                                        & 35.00   & 42.40                                                  & \multicolumn{1}{c|}{51.30}                                                         & 74.46                                                       & \textbf{83.29} \\ \midrule
\multicolumn{9}{c}{Augmented}                                                                                                                                                                                                                                                                                                                         \\ \midrule
SS     & 91.40  & 87.20    & \multicolumn{1}{c|}{76.10}                                                        & 86.50   & 80.90                                                  & \multicolumn{1}{c|}{82.50}                                                         & 91.32                                                       & \textbf{94.18} \\
SC     & 65.40  & 66.00    & \multicolumn{1}{c|}{72.70}                                                        & 72.10   & 76.90                                                  & \multicolumn{1}{c|}{68.40}                                                         & 80.61                                                       & \textbf{88.48} \\
CS     & 78.10  & 69.20    & \multicolumn{1}{c|}{48.10}                                                        & 61.60   & 47.30                                                  & \multicolumn{1}{c|}{68.40}                                                         & \textbf{81.20}                                              & 80.45          \\
CC     & 34.20  & 44.70    & \multicolumn{1}{c|}{52.60}                                                        & 36.80   & 55.20                                                  & \multicolumn{1}{c|}{47.30}                                                         & 65.79                                                       & \textbf{71.05} \\ \bottomrule
\end{tabular}%
\vspace{-0.75em}
\caption{Model accuracy across different categories. \textit{S} denotes '\textit{S}imple' and \textit{C} denotes '\textit{C}omplex'. The first and second letter represents chart and question type respectively.}
\vspace{-1.75em}
\label{tab:initial-score}
\end{table}

\paragraph{$(Q1)$ Does any model excel across all categories?} 
While no single model dominates all categories, GPT-4o and Gemini 1.5 Flash consistently demonstrate impressive performance, with GPT-4o leading in most cases. Among open-source models, InternLM stands out as the top performer. 

Models specifically trained on chart reasoning tasks (MatCha, UniChart) show exceptional performance while answering augmented questions, as highlighted in previous work as well (\cite{liu-etal-2023-matcha, masry-etal-2023-unichart}. This likely stems from their exposure to similar question formats during training, which is particularly evident in simple questions from the augmented set, where MatCha achieves a high accuracy of 91.40\%, followed by UniChart at 87.20\%.   However, they struggle significantly with reasoning-based questions, achieving as low as 25\% accuracy for complex chart and complex question pairs, highlighting the need for enhancement in the reasoning abilities of such models.

\paragraph{$(Q2)$ How do models perform across various chart types?}
Across all models, a consistent trend emerged: performance was consistently better on simple charts compared to complex charts, while comparing with the same question type. This behavior is likely attributable to the inherent difficulty in understanding and extracting values from complex charts. Factors like overlapping data points and the requirement of precise color resolution contributes to challenges in data extraction, increasing the difficulty of reasoning on such charts.

\paragraph{$(Q3)$ How do models perform across various question types?}
For the same chart type, models consistently perform better on simple questions compared to complex questions. This significant difference in scores highlights the limitations of certain models in fine-grained data extraction and reasoning.  GPT-4o and Gemini 1.5 Flash exhibit the smallest decrease in scores, indicating strong reasoning capabilities along with commendable data extraction skills. Smaller models, particularly those specifically trained on charts, struggle with questions requiring mathematical reasoning, despite their competence in basic data extraction. 

\paragraph{$(Q4)$ Do models struggle more with complex charts or complex questions?}
To further assess model capabilities, we compared performances of models on two categories: "Simple Charts, Complex Questions" and "Complex Charts, Simple Questions." This analysis reveals whether a model excels at visual data extraction (complex charts) or reasoning (complex questions).

Our results show that MLLMs like GPT-4o demonstrate strong reasoning skills, excelling in answering complex  questions. Gemini 1.5 Flash on the other hand performs consistently across both categories. Generalist and chart-based VLMs tend to favor the complex chart, simple question pair over the simple chart, complex question pair, suggesting limitations in reasoning abilities. This insight allows for targeted model fine-tuning to enhance specific domains where they lack dexterity. 

\begin{figure*}[!htbp]
    \centering
    \includegraphics[width=0.95\linewidth]{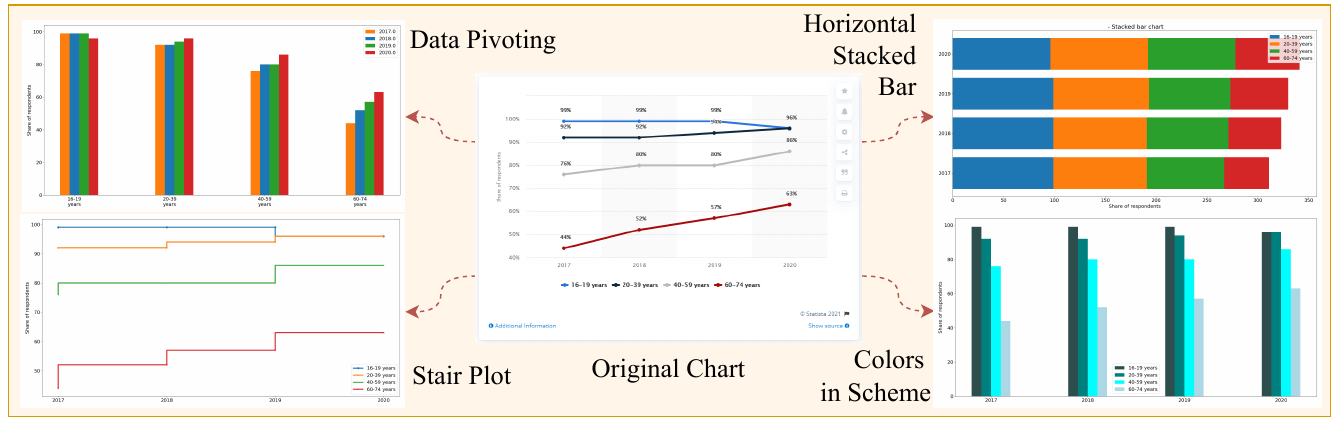}
    \vspace{-0.5em}
    \caption{\small Examples of different types of perturbations on the same original chart and data.}
    \vspace{-1.0em}
    \label{fig:perturbations}
\end{figure*}

\paragraph{$(Q5)$ Are there charts and questions where all models consistently fail to answer accurately?}\label{Q5} 
We focused on identifying patterns of model failure across all categories. In total, we found that 181 questions could not be answered by any model that we tested on. Given below are a few recurring difficulties that models faced:

\textbf{36/181 - Questions involving counting:} Models consistently struggled to accurately count objects when the number exceeded ten.

\textbf{30/181 - Charts containing similar colours:} Models struggled with charts which required discrimination between highly similar colors or shades of the same color.

\textbf{17/181 - Identifying colours from name:} Models struggle to accurately identify the color of chart elements when prompted to do so.

\textbf{17/181 - Charts involving summary statistics:} 
Models struggle to interpret charts with summary statistics, often confusing presented values with the need for recalculation. For example, given a chart of "Average of company revenues," they struggle to answer questions about "company A's average revenue," unsure whether to extract the value or recalculate it. This highlights a key limitation in their understanding of statistical representations.

\textbf{7/181- Tight pie charts:} In some instances, models incorrectly assigned labels to categories in pie charts with narrow slices, hence failing to identify the correct association.

A more detailed analysis on this topic has been presented in the \nameref{sec:appendix}.

\paragraph{$(Q6)$ How well do the models attend to the provided image for reasoning?}
To investigate the extent to which models rely on visual information versus their internal knowledge base, we conducted an experiment using \emph{blank images} and \emph{irrelevant charts}. We sampled 100 questions from each category and tested the top-performing models on their reasoning skills. 

Surprisingly, even when models were presented with irrelevant or blank images, some models successfully answered the questions, indicating a reliance on their pre-existing knowledge. This observation suggests potential leaks in testing data. Models were able to provide the correct answer even if the answer was factually incorrect from a real-world point-of-view, highlighting the need for masked evaluation sets for visual reasoning tasks.

\begin{table}[!htbp]
\centering
\small
\setlength{\tabcolsep}{3pt}
\begin{tabular}{@{}lcccc|cccc@{}}
\toprule
Model               & \multicolumn{4}{c|}{Blank Charts}                    & \multicolumn{4}{c}{Irrelevant Charts}               \\ \midrule
                    & SS          & SC         & CS          & CC          & SS         & SC         & CS          & CC          \\ \cmidrule(l){2-9} 
Gemini 1.5 Flash    & 0           & 0          & 0           & 0           & 0          & 2          & 2           & 4           \\
GPT-4o              & 0           & 3          & 0           & 3           & 0          & 2          & 1           & 6           \\
InternLM-XComposer2 & 2           & 3          & 8           & 6           & 1          & 5          & 3           & 2           \\
CogAgent-VQA        & \textbf{11} & 5          & 13          & 9           & 5          & 7          & \textbf{20} & 8           \\
Qwen-VL             & 7           & \textbf{9} & \textbf{21} & \textbf{17} & \textbf{9} & \textbf{8} & 13          & \textbf{14} \\ \bottomrule
\end{tabular}%
\vspace{-0.75em}
\caption{\small Performance of models when probed with blank and irrelevant charts.
\textit{S} denotes '\textit{S}imple' and \textit{C} denotes '\textit{C}omplex'. The first letter represents chart type and the second letter represents question type.}
\vspace{-0.75em}
\label{tab:probe}
\end{table}

Our analysis, detailed in Table \ref{tab:probe}, reveals that even large models like Gemini 1.5 Flash and GPT-4o were capable of answering few questions based on irrelevant charts, highlighting the needs of developing models that integrate visual information for robust visual reasoning capabilities. 

While our analysis reveals that models face challenges with certain categories of questions and charts, it also underscores the significant progress achieved in chart question answering (CQA) performance across various models.

\section{Are models robust on CQA?}
Another crucial aspect of our analysis involves investigating the robustness and consistency of these models across different visual representations of the same underlying data. Through the help of this probing, we aim to understand if model performance remains stable when presented with variations in chart types, styles, or aesthetics while conveying the same information.

Figure \ref{fig:perturbations} illustrates how an original chart can be converted into stair plots, bar charts, stacked representations, and more. These variations may differ in color schemes, patterns, legend positioning, and other chart-specific details like legend orientation and grid sizes on the x-axis and y-axis. Examining these variations can offer deeper insights into the data and improve the clarity of the visualizations.

\begin{table*}
\centering
\small
\setlength{\tabcolsep}{0.9pt}
\begin{tabular}{@{}lccccc|ccccc@{}}
\toprule
& \multicolumn{5}{c|}{Simple Questions}                                                                                                                                                                                                   & \multicolumn{5}{c}{Complex Questions}                                                                                                                                                                                                            \\ \hline
\multicolumn{1}{l}{}                                 & \multicolumn{2}{|c|}{MLLMs}                                                                     & \multicolumn{3}{c|}{Generalist VLMs}                                                                                                   & \multicolumn{2}{c|}{MLLMs}                                                                              & \multicolumn{3}{c}{Generalist VLMs}                                                                                                    \\ \hline
\multicolumn{1}{c|}{Category}                        & \begin{tabular}[c]{@{}c@{}}Gemini \\ 1.5 Flash\end{tabular} & \multicolumn{1}{c|}{GPT-4o}      & \begin{tabular}[c]{@{}c@{}} Qwen \\ VL \end{tabular}    & \begin{tabular}[c]{@{}c@{}}CogAgent\\ VQA\end{tabular} & \begin{tabular}[c]{@{}c@{}}InternLM\\ XComposer2\end{tabular} & \begin{tabular}[c]{@{}c@{}}Gemini \\ 1.5 Flash\end{tabular} & \multicolumn{1}{c|}{GPT-4o}               & \begin{tabular}[c]{@{}c@{}} Qwen \\ VL \end{tabular}    & \begin{tabular}[c]{@{}c@{}}CogAgent\\ VQA\end{tabular} & \begin{tabular}[c]{@{}c@{}}InternLM\\ XComposer2\end{tabular} \\ \hline
\multicolumn{1}{c|}{\textit{original\_chart}}        & \textit{\textbf{94}}                                        & \multicolumn{1}{c|}{\textit{89}} & \textit{62} & \textit{60}                                             & \textit{77}                                                    & \textit{71}                                                 & \multicolumn{1}{c|}{\textit{\textbf{82}}} & \textit{36} & \textit{45}                                             & \textit{49}                                                    \\ \hline
\multicolumn{1}{c|}{annotations}                     & 86                                                          & \multicolumn{1}{c|}{\textbf{90}} & 34          & 37                                                      & 59                                                             & 61                                                          & \multicolumn{1}{c|}{\textbf{77}}          & 31          & 40                                                      & 43                                                             \\
\multicolumn{1}{c|}{annotated\_bars}                 & 83                                                          & \multicolumn{1}{c|}{\textbf{89}} & 35          & 31                                                      & 64                                                             & 68                                                          & \multicolumn{1}{c|}{\textbf{74}}          & 26          & 28                                                      & 38                                                             \\
\multicolumn{1}{c|}{basic}                           & \textbf{67}                                                 & \multicolumn{1}{c|}{43}          & 17          & 17                                                      & 51                                                             & \textbf{55}                                                 & \multicolumn{1}{c|}{46}                   & 28          & 31                                                      & 37                                                             \\
\multicolumn{1}{c|}{color\_random}                   & \textbf{66}                                                 & \multicolumn{1}{c|}{45}          & 15          & 14                                                      & 51                                                             & \textbf{53}                                                 & \multicolumn{1}{c|}{49}                   & 21          & 29                                                      & 31                                                             \\
\multicolumn{1}{c|}{color\_scheme}                   & \textbf{56}                                                 & \multicolumn{1}{c|}{45}          & 16          & 13                                                      & 47                                                             & \textbf{55}                                                 & \multicolumn{1}{c|}{52}                   & 26          & 31                                                      & 40                                                             \\
\multicolumn{1}{c|}{data\_pivot}                     & \textbf{56}                                                 & \multicolumn{1}{c|}{43}          & 11          & 9                                                       & 46                                                             & \textbf{44}                                                 & \multicolumn{1}{c|}{23}                   & 28          & 27                                                      & 38                                                             \\
\multicolumn{1}{c|}{font}                            & \textbf{67}                                                 & \multicolumn{1}{c|}{49}          & 16          & 18                                                      & 34                                                             & \textbf{51}                                                 & \multicolumn{1}{c|}{43}                   & 26          & 28                                                      & 33                                                             \\
\multicolumn{1}{c|}{grid}                            & \textbf{67}                                                 & \multicolumn{1}{c|}{48}          & 18          & 16                                                      & 51                                                             & \textbf{52}                                                 & \multicolumn{1}{c|}{51}                   & 21          & 24                                                      & 34                                                             \\
\multicolumn{1}{c|}{hatching}                        & \textbf{57}                                                 & \multicolumn{1}{c|}{37}          & 11          & 9                                                       & 42                                                             & \textbf{49}                                                 & \multicolumn{1}{c|}{42}                   & 28          & 29                                                      & 37                                                             \\
\multicolumn{1}{c|}{horizontal\_grouped}             & \textbf{60}                                                 & \multicolumn{1}{c|}{32}          & 19          & 14                                                      & 49                                                             & \textbf{51}                                                 & \multicolumn{1}{c|}{42}                   & 29          & 29                                                      & 40                                                             \\
\multicolumn{1}{c|}{horizontal\_stacked}             & \textbf{30}                                                 & \multicolumn{1}{c|}{20}          & 16          & 11                                                      & 22                                                             & \textbf{59}                                                 & \multicolumn{1}{c|}{46}                   & 22          & 32                                                      & 43                                                             \\
\multicolumn{1}{c|}{legend\_position}                & \textbf{52}                                                 & \multicolumn{1}{c|}{44}          & 15          & 19                                                      & 49                                                             & \textbf{47}                                                 & \multicolumn{1}{c|}{46}                   & 28          & 30                                                      & 28                                                             \\
\multicolumn{1}{c|}{line\_representation}            & \textbf{52}                                                 & \multicolumn{1}{c|}{44}          & 13          & 18                                                      & 35                                                             & \textbf{42}                                                 & \multicolumn{1}{c|}{40}                   & 29          & 27                                                      & 33                                                             \\
\multicolumn{1}{c|}{log\_scale}                      & \textbf{38}                                                 & \multicolumn{1}{c|}{41}          & 11          & 9                                                       & 5                                                              & \textbf{55}                                                 & \multicolumn{1}{c|}{45}                   & 27          & 30                                                      & 38                                                             \\
\multicolumn{1}{c|}{only\_data\_color\_scheme}       & \textbf{62}                                                 & \multicolumn{1}{c|}{44}          & 17          & 18                                                      & 53                                                             & \textbf{51}                                                 & \multicolumn{1}{c|}{50}                   & 25          & 27                                                      & 39                                                             \\
\multicolumn{1}{c|}{replacing\_legend\_with\_labels} & \textbf{59}                                                 & \multicolumn{1}{c|}{48}          & 19          & 14                                                      & 41                                                             & 45                                                          & \multicolumn{1}{c|}{\textbf{56}}          & 30          & 28                                                      & 31                                                             \\
\multicolumn{1}{c|}{scaling\_size}                   & \textbf{63}                                                 & \multicolumn{1}{c|}{43}          & 11          & 13                                                      & 31                                                             & \textbf{47}                                                 & \multicolumn{1}{c|}{41}                   & 30          & 25                                                      & 28                                                             \\
\multicolumn{1}{c|}{scatter\_representation}         & \textbf{43}                                                 & \multicolumn{1}{c|}{38}          & 12          & 14                                                      & 37                                                             & \textbf{45}                                                 & \multicolumn{1}{c|}{44}                   & 23          & 27                                                      & 29                                                             \\
\multicolumn{1}{c|}{stacked}                         & \textbf{36}                                                 & \multicolumn{1}{c|}{28}          & 14          & 13                                                      & \textbf{36}                                                    & \textbf{47}                                                 & \multicolumn{1}{c|}{38}                   & 26          & 33                                                      & 32                                                             \\
\multicolumn{1}{c|}{stacked\_area}                   & \textbf{34}                                                 & \multicolumn{1}{c|}{24}          & 19          & 13                                                      & 31                                                             & \textbf{45}                                                 & \multicolumn{1}{c|}{41}                   & 26          & 34                                                      & 34                                                             \\
\multicolumn{1}{c|}{stair\_plot\_normal}             & \textbf{49}                                                 & \multicolumn{1}{c|}{41}          & 14          & 16                                                      & 41                                                             & \textbf{52}                                                 & \multicolumn{1}{c|}{43}                   & 17          & 29                                                      & 20                                                             \\
\multicolumn{1}{c|}{stair\_plot\_with\_marker}       & \textbf{55}                                                 & \multicolumn{1}{c|}{48}          & 13          & 20                                                      & 43                                                             & \textbf{57}                                                 & \multicolumn{1}{c|}{51}                   & 24          & 27                                                      & 45                                                             \\
\multicolumn{1}{c|}{stem\_plot}                      & 47                                                          & \multicolumn{1}{c|}{36}          & 12          & 12                                                      & \textbf{52}                                                    & \textbf{55}                                                 & \multicolumn{1}{c|}{41}                   & 22          & 28                                                      & 35                                                             \\
\multicolumn{1}{c|}{tick\_orientation}               & \textbf{66}                                                 & \multicolumn{1}{c|}{51}          & 19          & 14                                                      & 43                                                             & \textbf{42}                                                 & \multicolumn{1}{c|}{33}                   & 29          & 27                                                      & 27                                                             \\
\multicolumn{1}{c|}{tick\_position}                  & \textbf{56}                                                 & \multicolumn{1}{c|}{48}          & 21          & 16                                                      & 47                                                             & \textbf{49}                                                 & \multicolumn{1}{c|}{42}                   & 30          & 31                                                      & 30                                                             \\ \hline
\end{tabular}%
\vspace{-0.5em}
\caption{Model Performance on various perturbations on Complex Charts}
\vspace{-1.5em}
\label{tab:pert-complex}
\end{table*}

\subsection{Our RobustCQA Dataset} 
Following the initial dataset preparation, a perturbation dataset was created to rigorously assess the robustness of the top-performing models across diverse chart variations. We refer to this dataset as the RobustCQA dataset, which systematically manipulates various chart elements while preserving the underlying data. 

\vspace{-0.25em}
\paragraph{Creation} We identified 75 unique perturbation types for both simple and complex charts. These perturbations cover a broad spectrum of visual variations, including:
\begin{itemize}
\setlength\itemsep{0.0em}
    \item \textbf{Color Scheme Changes:} Modifying color palettes, gradients and hues.
    \item \textbf{Chart Type Variations:} Experimenting with line plots, bar plots, stair plots, stem plots and other less commonly used chart types.
    \item \textbf{Legend and Axis Modification:} Altering label position, formatting, and positioning of legend and axis elements.
\end{itemize}

The perturbed charts were generated using the Matplotlib library. We ensured that \textbf{only one element is altered per perturbation} keeping the rest of the elements the same. The tables from the ChartQA dataset served as the source for the underlying data.

\paragraph{Human Verification} To ensure the quality and relevance of our dataset, a rigorous manual annotation process was employed. Expert evaluators meticulously verified each perturbed chart, assessing how easily comprehensible and answerable each perturbed chart was. They also evaluated the relevance of each perturbation to the specific chart type, refining the perturbation set to include only meaningful variations. The underlying tables were also thoroughly verified to confirm that the generated questions remained answerable based on the chart data.  This comprehensive evaluation was facilitated by a custom-built annotation platform, specifically designed to streamline the manual annotation process and ensure high-quality data. 

\paragraph{Final Dataset} The original 75 perturbations were then grouped into categories of related perturbations to create the final dataset. This set consists of 22 unique perturbation categories for simple charts and 25 such categories for complex charts, covering a wide range of visual variations. 

To ensure a fair analysis of model robustness across perturbations, 100 questions were sampled for each chart type and question type pair. This resulted in a total of 400 unique table and QA pairs for our final evaluation. This standardized question set allows for direct comparison of model performance across different visual representations. We finally compare the results of all perturbations against the basic or default Matplotlib chart. 

A detailed breakdown of the perturbation categories along with examples has been included in the \nameref{sec:appendix}. 

\subsection{Methodology} 
To delve deeper into the performance and limitations of leading chart question answering models, we evaluated Qwen-VL, CogAgent-VQA, InternLM-XComposer2 (open-source VLMs) and Gemini 1.5 Flash, GPT-4o (closed-source MLLMs) using our RobustCQA dataset. We employed a similar evaluation metric as described previously, leveraging an extractor LLM for smaller models to ensure consistent output format, and analyzed all models through Zero-Shot Chain-of-Thought prompting. 

\subsection{Results and Discussion}
The results obtained for perturbations on complex charts have been highlighted in table \ref{tab:pert-complex}. The results for perturbations on simple charts has been presented in table \ref{tab:pert-simple} in the \nameref{sec:appendix}. 

\paragraph{$(Q1)$ Does model performance stay consistent with perturbed charts?}
The results reveal a significant performance degradation for most models when confronted with perturbations. While performance generally decreases across all models, some exhibit more drastic drops. Among open sourced models, InternLM-XComposer2, and among closed source models, Gemini 1.5 Flash proved to be the most consistent across various perturbations. Open sourced models like CogAgent-VQA and Qwen-VL and even closed source models like GPT-4o displayed relatively low accuracy with most perturbations, highlighting a potential lack of robust data extraction skills. Manual analysis of the responses from the models highlighted the importance of improving data extraction for non-annotated charts to enhance model robustness in chart-based tasks. 

\paragraph{$(Q2)$ Are there specific perturbations that help enhance model performance?}
Our experiments highlighted several perturbations that improved model performance. Across all models, annotated data points consistently boosted accuracy. While the most beneficial plot type varied across models and question/chart categories, annotated bar graphs emerged as a consistently positive influence. 

In addition to that, grids to act as reference points for data extraction and better tick orientation also contributed positively. Furthermore, labelling the lines to reduce the complexity of color resolution and placing the legend optimally to ensure that it doesn't obscure crucial data points also helped the models. We also noticed that increasing the font size played an important role in aiding all models, especially smaller models. The results for the same have been presented in Table \ref{tab:font-tale}.

\paragraph{$(Q3)$ Are there perturbations which are always detrimental to the model performance?}
Our analysis reveals that while models demonstrate promising performance on standard chart datasets, they struggle with robustness when faced with visual perturbations. While annotations generally help improve model performance, most other perturbations negatively impacted model accuracy. 

Notable ones among them include logarithmic scales which can be challenging for humans as well. Additionally, models also struggle significantly with horizontal chart variations, particularly horizontal stacked charts. In general it was noticed that models struggled to reason on stacked charts, possibly due to the requirement of additional mathematical reasoning for data extraction. Stair plots also caused significant trouble as models could not identify the precise data point to refer to. Our findings emphasize the need of more diverse datasets along with more robust models that can effectively interpret visual information beyond just simple visual cues.

\paragraph{$(Q4)$ Are there certain perturbations which are more effective for certain question types?}
Our analysis suggests chart type effectiveness varies by question type. For instance, line charts help in visualizing trends and correlations. Stacked bar charts are generally unsuitable except for questions that require data aggregation. Bar charts, while useful for comparing individual values within a certain group, prove to not be good for showcasing correlations across different groups or entities. Our analysis helps with understanding and creating suitable charts for domain specific tasks.

\paragraph{$(Q5)$ Does the effect of each perturbation type vary across models?}
The impact of each perturbation on model performance exhibits significant variation. While question and chart type play a role, for a given model, certain perturbations consistently prove more helpful or harmful. This nuanced effect of perturbation type on the model performance is detailed in Table \ref{tab:best5}, \ref{tab:worst5}. We believe that our analysis allows us to identify specific areas for helping improve each model through targeted model fine-tuning.

Additional insights and details obtained from our experiments have been presented in the \nameref{sec:appendix}.

\section{Related Work}
Chart comprehension and question answering (CQA) are critical domains with a growing body of research. While existing CQA datasets assess models' advanced reasoning capabilities over charts, many face significant limitations. These include small dataset sizes \cite{visqa2020}, reliance on template-based questions and synthetically generated charts \cite{Methani_2020_WACV, chaudhry2020leaf, kafle2018dvqa, han2023chartllama}, restriction to specific domains \cite{Methani_2020_WACV, ahmed2023realcqa, li2023scigraphqalargescalesyntheticmultiturn}, or focusing solely on open-domain question answering \cite{kantharaj-etal-2022-opencqa}. Even the current state-of-the-art dataset, Chart QA \cite{masry-etal-2022-chartqa}, has limitations due to the lack of classification labels for more meaningful analysis, and limited variation in chart types.

More recent datasets, such as ChartX \cite{xia2024chartx}, have expanded the range of chart types analyzed. ChartBench \cite{xu2024chartbench} and MMC \cite{liu-etal-2024-mmc} focus on large-scale datasets with more diverse chart types.

A very recent work, CharXiv \cite{wang2024charxivchartinggapsrealistic}, provides extensive evaluations across a range of charts and questions, including both reasoning-based and descriptive queries. They also perform ablation studies by modifying charts and questions.

However, to the best of our knowledge, RobustCQA is the first dataset to systematically perturb all elements within a chart, enabling fine-grained analysis of factors affecting model performance. Additionally, we conduct a detailed analysis of the Chart QA dataset based on question and chart complexity, which has not been done at this level of detail before.

\paragraph{Modeling approaches for charts} Various approaches have been developed for chart modeling. This includes models specifically designed for chart comprehension and reasoning, built with the end-to-end goal of reasoning over charts \cite{liu-etal-2023-matcha, masry-etal-2023-unichart, singh-shekhar-2020-stl}, as well as models that convert charts into intermediate table formats \cite{liu-etal-2023-deplot}, enabling reasoning by generalized large language models (LLMs) through Chain of Thought \cite{wei2022cot} or Program of Thought \cite{chen2023program} prompting. Additionally, there are generalized models used for multi-modal reasoning tasks, including chart comprehension \cite{team2023gemini, achiam2023gpt, Qwen-VL, internlmxcomposer2, hong2024cogagent}. Recent efforts have also focused on developing smaller, yet accurate models for this task \cite{wang2024charxivchartinggapsrealistic}.

While these approaches have shown significant progress, their specific failure points remain unclear. A recent study \cite{islam2024largevisionlanguagemodels} analyzed the performance of GPT-4v and Gemini, providing a broad evaluation of these models across various chart comprehension tasks, including question answering, summarization, and fact-checking. In contrast, our work focuses specifically on CQA across a broader range of models, offering an in-depth analysis of the question and chart types contributing to model failures. Our contribution identifies the exact question types and chart elements that lead to model errors, offering key insights to improve model performance.

\paragraph{Vision-Language Model Robustness} Recent studies have highlighted the vulnerability of models to attacks and perturbations \cite{ma2024orbust, zhao2023evaluate}, raising concerns about their robustness in real-world deployment. Motivated by this, we developed a robustness benchmark specifically for chart question answering (CQA). While previous work \cite{gupta2024enhancingquestionansweringcharts} analyzed models like DePlot and MatCha on perturbed charts, focusing on questions related to structural and visual context, our study examines general reasoning questions. This approach helps us assess how variations in the visual representation of the same data affect model performance.

\section{Conclusion} 
This research introduces ChartQA-Split and RobustCQA, the first datasets dedicated to understanding model consistency across complexities and robustness to visual perturbations in chart question answering. Our evaluation of SOTA models, including baselines and VLMs/MLLMs, using a zero-shot chain-of-thought setting, reveals significant challenges in both areas.  We perform an in-depth analysis of model weaknesses and identify key areas for improvement, such as enhancing data extraction for non-annotated charts and developing models that can effectively interpret complex visual information, taking every possible visual cue into consideration. Our work provides a foundation for future research in developing more robust and reliable chart question answering systems. 

\paragraph{Future Directions.}
Our perturbation analysis provides a nuanced understanding of model performance by revealing both universal and model-specific vulnerabilities and strengths. This insight drives targeted improvements: \textbf{Model Pretraining:} Focusing on perturbations that affect models allows for effective fine-tuning to address weaknesses. \textbf{Perturbation-Aware Training:} Integrating specific perturbations during training enhances overall robustness, helping models develop resilience against challenges. \textbf{Interpretable Models:} Understanding the impact of perturbations aids in debugging and building explainable models, fostering the development of reliable and transparent chart understanding and reasoning systems.

\section*{Limitations}
The presented work exhibits a few limitations. First, our data was obtained from a singular dataset, and we used only one plotting software for testing the perturbations. Expanding the dataset to include diverse sources and exploring various plotting libraries would strengthen the findings and improve generalizability. Second, the dataset is limited to English, while models are developed and evaluated on a wide variety of languages. Future research is required to expand the domain beyond English. Third, we were not able to cover a few chart types in the course of our analysis in order to make a more generalized perturbation set. This included pie and doughnut charts, pyramid and funnel charts as well as radar charts. Due to metadata limitations and the complexity of adapting data for chart representation, these charts were excluded. Fourth, inconsistent metadata of the original dataset sometimes lacked visual captions present in the original charts, which could not be replicated in the perturbed charts. Because of this, we were not able to identify attributes pertaining to chart elements, for example, colour. 

\section*{Ethics Statement}
This research adheres to the ACL code of ethics, acknowledging and addressing potential ethical implications.  While LLMs assisted in writing and presentation, all ideas and conclusions are solely attributed to the authors. The research promotes responsible and fair use of methodologies, ensuring transparency and reproducibility.  We plan to release all scripts, resources, comprehensive documentation, evaluation metrics, datasets, model specifications, and prompting methods to enable others to build upon our work. We strive to present our findings clearly and accurately, avoiding exaggerated claims or misinterpretations.

\section*{Acknowledgement}
Research was sponsored by the Army Research Office and was accomplished under Grant Number W911NF-20-1-0080. The views and conclusions contained in this document are those of the authors and should not be interpreted as representing the official policies, either expressed or implied, of the Army Research Office or the U.S. Government. The U.S. Government is authorized to reproduce and distribute reprints for Government purposes notwithstanding any copyright notation herein. This work was partially funded by ONR Contract N00014-23-1-2365. Lastly, we acknowledge the generous gift from Adobe.

\bibliography{main}

\begin{thebibliography}{34}
\expandafter\ifx\csname natexlab\endcsname\relax\def\natexlab#1{#1}\fi

\bibitem[{Achiam et~al.(2023)Achiam, Adler, Agarwal, Ahmad, Akkaya, Aleman, Almeida, Altenschmidt, Altman, Anadkat et~al.}]{achiam2023gpt}
Josh Achiam, Steven Adler, Sandhini Agarwal, Lama Ahmad, Ilge Akkaya, Florencia~Leoni Aleman, Diogo Almeida, Janko Altenschmidt, Sam Altman, Shyamal Anadkat, et~al. 2023.
\newblock Gpt-4 technical report.
\newblock \emph{arXiv preprint arXiv:2303.08774}.

\bibitem[{Ahmed et~al.(2023)Ahmed, Jawade, Pandey, Setlur, and Govindaraju}]{ahmed2023realcqa}
Saleem Ahmed, Bhavin Jawade, Shubham Pandey, Srirangaraj Setlur, and Venu Govindaraju. 2023.
\newblock Realcqa: Scientific chart question answering as a test-bed for first-order logic.
\newblock In \emph{International Conference on Document Analysis and Recognition}, pages 66--83. Springer.

\bibitem[{Bai et~al.(2023{\natexlab{a}})Bai, Bai, Chu, Cui, Dang, Deng, Fan, Ge, Han, Huang, Hui, Ji, Li, Lin, Lin, Liu, Liu, Lu, Lu, Ma, Men, Ren, Ren, Tan, Tan, Tu, Wang, Wang, Wang, Wu, Xu, Xu, Yang, Yang, Yang, Yang, Yao, Yu, Yuan, Yuan, Zhang, Zhang, Zhang, Zhang, Zhou, Zhou, Zhou, and Zhu}]{qwen}
Jinze Bai, Shuai Bai, Yunfei Chu, Zeyu Cui, Kai Dang, Xiaodong Deng, Yang Fan, Wenbin Ge, Yu~Han, Fei Huang, Binyuan Hui, Luo Ji, Mei Li, Junyang Lin, Runji Lin, Dayiheng Liu, Gao Liu, Chengqiang Lu, Keming Lu, Jianxin Ma, Rui Men, Xingzhang Ren, Xuancheng Ren, Chuanqi Tan, Sinan Tan, Jianhong Tu, Peng Wang, Shijie Wang, Wei Wang, Shengguang Wu, Benfeng Xu, Jin Xu, An~Yang, Hao Yang, Jian Yang, Shusheng Yang, Yang Yao, Bowen Yu, Hongyi Yuan, Zheng Yuan, Jianwei Zhang, Xingxuan Zhang, Yichang Zhang, Zhenru Zhang, Chang Zhou, Jingren Zhou, Xiaohuan Zhou, and Tianhang Zhu. 2023{\natexlab{a}}.
\newblock Qwen technical report.
\newblock \emph{arXiv preprint arXiv:2309.16609}.

\bibitem[{Bai et~al.(2023{\natexlab{b}})Bai, Bai, Yang, Wang, Tan, Wang, Lin, Zhou, and Zhou}]{Qwen-VL}
Jinze Bai, Shuai Bai, Shusheng Yang, Shijie Wang, Sinan Tan, Peng Wang, Junyang Lin, Chang Zhou, and Jingren Zhou. 2023{\natexlab{b}}.
\newblock Qwen-vl: A versatile vision-language model for understanding, localization, text reading, and beyond.
\newblock \emph{arXiv preprint arXiv:2308.12966}.

\bibitem[{Cai et~al.(2024)Cai, Cao, Chen, Chen, Chen, Chen, Chen, Chen, Chen, Chu et~al.}]{cai2024internlm2}
Zheng Cai, Maosong Cao, Haojiong Chen, Kai Chen, Keyu Chen, Xin Chen, Xun Chen, Zehui Chen, Zhi Chen, Pei Chu, et~al. 2024.
\newblock Internlm2 technical report.
\newblock \emph{CoRR}.

\bibitem[{Chaudhry et~al.(2020)Chaudhry, Shekhar, Gupta, Maneriker, Bansal, and Joshi}]{chaudhry2020leaf}
Ritwick Chaudhry, Sumit Shekhar, Utkarsh Gupta, Pranav Maneriker, Prann Bansal, and Ajay Joshi. 2020.
\newblock Leaf-qa: Locate, encode \& attend for figure question answering.
\newblock In \emph{2020 IEEE Winter Conference on Applications of Computer Vision (WACV)}. IEEE.

\bibitem[{Chen et~al.(2023)Chen, Ma, Wang, and Cohen}]{chen2023program}
Wenhu Chen, Xueguang Ma, Xinyi Wang, and William~W. Cohen. 2023.
\newblock \href {https://openreview.net/forum?id=YfZ4ZPt8zd} {Program of thoughts prompting: Disentangling computation from reasoning for numerical reasoning tasks}.
\newblock \emph{Transactions on Machine Learning Research}.

\bibitem[{Dong et~al.(2024)Dong, Zhang, Zang, Cao, Wang, Ouyang, Wei, Zhang, Duan, Cao, Zhang, Li, Yan, Gao, Zhang, Li, Li, Chen, He, Zhang, Qiao, Lin, and Wang}]{internlmxcomposer2}
Xiaoyi Dong, Pan Zhang, Yuhang Zang, Yuhang Cao, Bin Wang, Linke Ouyang, Xilin Wei, Songyang Zhang, Haodong Duan, Maosong Cao, Wenwei Zhang, Yining Li, Hang Yan, Yang Gao, Xinyue Zhang, Wei Li, Jingwen Li, Kai Chen, Conghui He, Xingcheng Zhang, Yu~Qiao, Dahua Lin, and Jiaqi Wang. 2024.
\newblock Internlm-xcomposer2: Mastering free-form text-image composition and comprehension in vision-language large model.
\newblock \emph{arXiv preprint arXiv:2401.16420}.

\bibitem[{Ghosh et~al.(2024)Ghosh, Acharya, Saha, Jain, and Chadha}]{ghosh2024exploringfrontiervisionlanguagemodels}
Akash Ghosh, Arkadeep Acharya, Sriparna Saha, Vinija Jain, and Aman Chadha. 2024.
\newblock \href {http://arxiv.org/abs/2404.07214} {Exploring the frontier of vision-language models: A survey of current methodologies and future directions}.

\bibitem[{Gupta et~al.(2024)Gupta, Gupta, Zhang, He, Zhang, and Shah}]{gupta2024enhancingquestionansweringcharts}
Ashim Gupta, Vivek Gupta, Shuo Zhang, Yujie He, Ning Zhang, and Shalin Shah. 2024.
\newblock \href {http://arxiv.org/abs/2406.10085} {Enhancing question answering on charts through effective pre-training tasks}.

\bibitem[{Han et~al.(2023)Han, Zhang, Chen, Yang, Wang, Yu, Fu, and Zhang}]{han2023chartllama}
Yucheng Han, Chi Zhang, Xin Chen, Xu~Yang, Zhibin Wang, Gang Yu, Bin Fu, and Hanwang Zhang. 2023.
\newblock \href {http://arxiv.org/abs/2311.16483} {Chartllama: A multimodal llm for chart understanding and generation}.

\bibitem[{Hong et~al.(2024)Hong, Wang, Lv, Xu, Yu, Ji, Wang, Wang, Dong, Ding et~al.}]{hong2024cogagent}
Wenyi Hong, Weihan Wang, Qingsong Lv, Jiazheng Xu, Wenmeng Yu, Junhui Ji, Yan Wang, Zihan Wang, Yuxiao Dong, Ming Ding, et~al. 2024.
\newblock Cogagent: A visual language model for gui agents.
\newblock In \emph{Proceedings of the IEEE/CVF Conference on Computer Vision and Pattern Recognition}, pages 14281--14290.

\bibitem[{Islam et~al.(2024)Islam, Rahman, Masry, Laskar, Nayeem, and Hoque}]{islam2024largevisionlanguagemodels}
Mohammed~Saidul Islam, Raian Rahman, Ahmed Masry, Md~Tahmid~Rahman Laskar, Mir~Tafseer Nayeem, and Enamul Hoque. 2024.
\newblock \href {http://arxiv.org/abs/2406.00257} {Are large vision language models up to the challenge of chart comprehension and reasoning? an extensive investigation into the capabilities and limitations of lvlms}.

\bibitem[{Kafle et~al.(2018)Kafle, Price, Cohen, and Kanan}]{kafle2018dvqa}
Kushal Kafle, Brian Price, Scott Cohen, and Christopher Kanan. 2018.
\newblock Dvqa: Understanding data visualizations via question answering.
\newblock In \emph{Proceedings of the IEEE conference on computer vision and pattern recognition}, pages 5648--5656.

\bibitem[{Kantharaj et~al.(2022)Kantharaj, Do, Leong, Tan, Hoque, and Joty}]{kantharaj-etal-2022-opencqa}
Shankar Kantharaj, Xuan~Long Do, Rixie~Tiffany Leong, Jia~Qing Tan, Enamul Hoque, and Shafiq Joty. 2022.
\newblock \href {https://doi.org/10.18653/v1/2022.emnlp-main.811} {{O}pen{CQA}: Open-ended question answering with charts}.
\newblock In \emph{Proceedings of the 2022 Conference on Empirical Methods in Natural Language Processing}, pages 11817--11837, Abu Dhabi, United Arab Emirates. Association for Computational Linguistics.

\bibitem[{Kim et~al.(2020)Kim, Hoque, and Agrawala}]{visqa2020}
Dae~Hyun Kim, Enamul Hoque, and Maneesh Agrawala. 2020.
\newblock \href {https://doi.org/10.1145/3313831.3376467} {Answering questions about charts and generating visual explanations}.
\newblock In \emph{Proceedings of the 2020 CHI Conference on Human Factors in Computing Systems}, CHI '20, page 1–13, New York, NY, USA. Association for Computing Machinery.

\bibitem[{Lee et~al.(2023)Lee, Joshi, Turc, Hu, Liu, Eisenschlos, Khandelwal, Shaw, Chang, and Toutanova}]{lee2023pix2struct}
Kenton Lee, Mandar Joshi, Iulia~Raluca Turc, Hexiang Hu, Fangyu Liu, Julian~Martin Eisenschlos, Urvashi Khandelwal, Peter Shaw, Ming-Wei Chang, and Kristina Toutanova. 2023.
\newblock Pix2struct: Screenshot parsing as pretraining for visual language understanding.
\newblock In \emph{International Conference on Machine Learning}, pages 18893--18912. PMLR.

\bibitem[{Li and Tajbakhsh(2023)}]{li2023scigraphqalargescalesyntheticmultiturn}
Shengzhi Li and Nima Tajbakhsh. 2023.
\newblock \href {http://arxiv.org/abs/2308.03349} {Scigraphqa: A large-scale synthetic multi-turn question-answering dataset for scientific graphs}.

\bibitem[{Liu et~al.(2023{\natexlab{a}})Liu, Eisenschlos, Piccinno, Krichene, Pang, Lee, Joshi, Chen, Collier, and Altun}]{liu-etal-2023-deplot}
Fangyu Liu, Julian Eisenschlos, Francesco Piccinno, Syrine Krichene, Chenxi Pang, Kenton Lee, Mandar Joshi, Wenhu Chen, Nigel Collier, and Yasemin Altun. 2023{\natexlab{a}}.
\newblock \href {https://doi.org/10.18653/v1/2023.findings-acl.660} {{D}e{P}lot: One-shot visual language reasoning by plot-to-table translation}.
\newblock In \emph{Findings of the Association for Computational Linguistics: ACL 2023}, pages 10381--10399, Toronto, Canada. Association for Computational Linguistics.

\bibitem[{Liu et~al.(2023{\natexlab{b}})Liu, Piccinno, Krichene, Pang, Lee, Joshi, Altun, Collier, and Eisenschlos}]{liu-etal-2023-matcha}
Fangyu Liu, Francesco Piccinno, Syrine Krichene, Chenxi Pang, Kenton Lee, Mandar Joshi, Yasemin Altun, Nigel Collier, and Julian Eisenschlos. 2023{\natexlab{b}}.
\newblock \href {https://doi.org/10.18653/v1/2023.acl-long.714} {{M}at{C}ha: Enhancing visual language pretraining with math reasoning and chart derendering}.
\newblock In \emph{Proceedings of the 61st Annual Meeting of the Association for Computational Linguistics (Volume 1: Long Papers)}, pages 12756--12770, Toronto, Canada. Association for Computational Linguistics.

\bibitem[{Liu et~al.(2024)Liu, Wang, Yao, Chen, Song, Cho, Yacoob, and Yu}]{liu-etal-2024-mmc}
Fuxiao Liu, Xiaoyang Wang, Wenlin Yao, Jianshu Chen, Kaiqiang Song, Sangwoo Cho, Yaser Yacoob, and Dong Yu. 2024.
\newblock \href {https://aclanthology.org/2024.naacl-long.70} {{MMC}: Advancing multimodal chart understanding with large-scale instruction tuning}.
\newblock In \emph{Proceedings of the 2024 Conference of the North American Chapter of the Association for Computational Linguistics: Human Language Technologies (Volume 1: Long Papers)}, pages 1287--1310, Mexico City, Mexico. Association for Computational Linguistics.

\bibitem[{Ma et~al.(2024)Ma, Wang, Kong, Wang, Liu, Pei, and Zhao}]{ma2024orbust}
J.~Ma, P.~Wang, D.~Kong, Z.~Wang, J.~Liu, H.~Pei, and J.~Zhao. 2024.
\newblock \href {https://doi.org/10.1109/TPAMI.2024.3366154} {Robust visual question answering: Datasets, methods, and future challenges}.
\newblock \emph{IEEE Transactions on Pattern Analysis I\&; Machine Intelligence}, 46(08):5575--5594.

\bibitem[{Masry et~al.(2022)Masry, Do, Tan, Joty, and Hoque}]{masry-etal-2022-chartqa}
Ahmed Masry, Xuan~Long Do, Jia~Qing Tan, Shafiq Joty, and Enamul Hoque. 2022.
\newblock \href {https://doi.org/10.18653/v1/2022.findings-acl.177} {{C}hart{QA}: A benchmark for question answering about charts with visual and logical reasoning}.
\newblock In \emph{Findings of the Association for Computational Linguistics: ACL 2022}, pages 2263--2279, Dublin, Ireland. Association for Computational Linguistics.

\bibitem[{Masry et~al.(2023)Masry, Kavehzadeh, Do, Hoque, and Joty}]{masry-etal-2023-unichart}
Ahmed Masry, Parsa Kavehzadeh, Xuan~Long Do, Enamul Hoque, and Shafiq Joty. 2023.
\newblock \href {https://doi.org/10.18653/v1/2023.emnlp-main.906} {{U}ni{C}hart: A universal vision-language pretrained model for chart comprehension and reasoning}.
\newblock In \emph{Proceedings of the 2023 Conference on Empirical Methods in Natural Language Processing}, pages 14662--14684, Singapore. Association for Computational Linguistics.

\bibitem[{Meng et~al.(2024)Meng, Shao, Lu, Gao, Zhang, Qiao, and Luo}]{meng2024chartassisstant}
Fanqing Meng, Wenqi Shao, Quanfeng Lu, Peng Gao, Kaipeng Zhang, Yu~Qiao, and Ping Luo. 2024.
\newblock Chartassisstant: A universal chart multimodal language model via chart-to-table pre-training and multitask instruction tuning.
\newblock \emph{arXiv preprint arXiv:2401.02384}.

\bibitem[{Methani et~al.(2020)Methani, Ganguly, Khapra, and Kumar}]{Methani_2020_WACV}
Nitesh Methani, Pritha Ganguly, Mitesh~M. Khapra, and Pratyush Kumar. 2020.
\newblock Plotqa: Reasoning over scientific plots.
\newblock In \emph{The IEEE Winter Conference on Applications of Computer Vision (WACV)}.

\bibitem[{Singh and Shekhar(2020)}]{singh-shekhar-2020-stl}
Hrituraj Singh and Sumit Shekhar. 2020.
\newblock \href {https://doi.org/10.18653/v1/2020.emnlp-main.264} {{STL-CQA}: Structure-based transformers with localization and encoding for chart question answering}.
\newblock In \emph{Proceedings of the 2020 Conference on Empirical Methods in Natural Language Processing (EMNLP)}, pages 3275--3284, Online. Association for Computational Linguistics.

\bibitem[{Tay et~al.(2022)Tay, Dehghani, Tran, Garc{\'i}a, Wei, Wang, Chung, Bahri, Schuster, Zheng, Zhou, Houlsby, and Metzler}]{Tay2022UL2UL}
Yi~Tay, Mostafa Dehghani, Vinh~Q. Tran, Xavier Garc{\'i}a, Jason Wei, Xuezhi Wang, Hyung~Won Chung, Dara Bahri, Tal Schuster, Huaixiu~Steven Zheng, Denny Zhou, Neil Houlsby, and Donald Metzler. 2022.
\newblock \href {https://api.semanticscholar.org/CorpusID:252780443} {Ul2: Unifying language learning paradigms}.
\newblock In \emph{International Conference on Learning Representations}.

\bibitem[{Team et~al.(2023)Team, Anil, Borgeaud, Wu, Alayrac, Yu, Soricut, Schalkwyk, Dai, Hauth et~al.}]{team2023gemini}
Gemini Team, Rohan Anil, Sebastian Borgeaud, Yonghui Wu, Jean-Baptiste Alayrac, Jiahui Yu, Radu Soricut, Johan Schalkwyk, Andrew~M Dai, Anja Hauth, et~al. 2023.
\newblock Gemini: a family of highly capable multimodal models.
\newblock \emph{arXiv preprint arXiv:2312.11805}.

\bibitem[{Wang et~al.(2024)Wang, Xia, He, Chen, Liu, Zhu, Liang, Wu, Liu, Malladi, Chevalier, Arora, and Chen}]{wang2024charxivchartinggapsrealistic}
Zirui Wang, Mengzhou Xia, Luxi He, Howard Chen, Yitao Liu, Richard Zhu, Kaiqu Liang, Xindi Wu, Haotian Liu, Sadhika Malladi, Alexis Chevalier, Sanjeev Arora, and Danqi Chen. 2024.
\newblock \href {http://arxiv.org/abs/2406.18521} {Charxiv: Charting gaps in realistic chart understanding in multimodal llms}.

\bibitem[{Wei et~al.(2022)Wei, Wang, Schuurmans, Bosma, ichter, Xia, Chi, Le, and Zhou}]{wei2022cot}
Jason Wei, Xuezhi Wang, Dale Schuurmans, Maarten Bosma, brian ichter, Fei Xia, Ed~Chi, Quoc~V Le, and Denny Zhou. 2022.
\newblock \href {https://proceedings.neurips.cc/paper_files/paper/2022/file/9d5609613524ecf4f15af0f7b31abca4-Paper-Conference.pdf} {Chain-of-thought prompting elicits reasoning in large language models}.
\newblock In \emph{Advances in Neural Information Processing Systems}, volume~35, pages 24824--24837. Curran Associates, Inc.

\bibitem[{Xia et~al.(2024)Xia, Zhang, Ye, Yan, Liu, Zhou, Chen, Dou, Shi, Yan et~al.}]{xia2024chartx}
Renqiu Xia, Bo~Zhang, Hancheng Ye, Xiangchao Yan, Qi~Liu, Hongbin Zhou, Zijun Chen, Min Dou, Botian Shi, Junchi Yan, et~al. 2024.
\newblock Chartx \& chartvlm: A versatile benchmark and foundation model for complicated chart reasoning.
\newblock \emph{arXiv preprint arXiv:2402.12185}.

\bibitem[{Xu et~al.(2024)Xu, Du, Qi, Xu, Yuan, and Guo}]{xu2024chartbench}
Zhengzhuo Xu, Sinan Du, Yiyan Qi, Chengjin Xu, Chun Yuan, and Jian Guo. 2024.
\newblock \href {http://arxiv.org/abs/2312.15915} {Chartbench: A benchmark for complex visual reasoning in charts}.

\bibitem[{Zhao et~al.(2023)Zhao, Pang, Du, Yang, Li, Cheung, and Lin}]{zhao2023evaluate}
Yunqing Zhao, Tianyu Pang, Chao Du, Xiao Yang, Chongxuan Li, Ngai-Man Cheung, and Min Lin. 2023.
\newblock On evaluating adversarial robustness of large vision-language models.
\newblock In \emph{Thirty-seventh Conference on Neural Information Processing Systems}.

\end{thebibliography}
\bibliographystyle{acl_natbib}

\section*{Appendix}
\appendix
\label{sec:appendix}

\paragraph{Effect of Font-size on Models} 
Table \ref{tab:font-tale} illustrates the significant impact of font size on model performance. Increasing the font size improves the OCR capabilities of visual language models (VLMs), suggesting that larger font sizes can enhance model accuracy in chart comprehension tasks. This finding indicates that adjusting font size could serve as an effective preprocessing step for boosting performance in such tasks.

\begin{table}[!ht]
\centering
\resizebox{\columnwidth}{!}{%
\begin{tabular}{@{}lcc|cc@{}}
\toprule
 & \multicolumn{2}{c|}{\textbf{Gemini 1.5 Flash}} & \multicolumn{2}{c}{\textbf{Qwen VL}} \\ \midrule
\multicolumn{1}{c|}{\multirow{2}{*}{\textbf{\begin{tabular}[c]{@{}c@{}}Perturbation \\ types\end{tabular}}}} & \multicolumn{1}{c|}{\multirow{2}{*}{\textbf{\begin{tabular}[c]{@{}c@{}}Small\\ Font\end{tabular}}}} & \multirow{2}{*}{\textbf{\begin{tabular}[c]{@{}c@{}}Big\\ Font\end{tabular}}} & \multicolumn{1}{c|}{\multirow{2}{*}{\textbf{\begin{tabular}[c]{@{}c@{}}Small\\ Font\end{tabular}}}} & \multirow{2}{*}{\textbf{\begin{tabular}[c]{@{}c@{}}Big\\ Font\end{tabular}}} \\
\multicolumn{1}{c|}{} & \multicolumn{1}{c|}{} &  & \multicolumn{1}{c|}{} &  \\ \midrule
\multicolumn{1}{l|}{Normal line plot} & \multicolumn{1}{c|}{62} & 63 & \multicolumn{1}{c|}{6} & 27 \\
\multicolumn{1}{l|}{\begin{tabular}[c]{@{}l@{}}Colors in a given \\ scheme (line)\end{tabular}} & \multicolumn{1}{c|}{56} & 63 & \multicolumn{1}{c|}{7} & 26 \\
\multicolumn{1}{l|}{Colors random (scatter)} & \multicolumn{1}{c|}{46} & 57 & \multicolumn{1}{c|}{5} & 21 \\
\multicolumn{1}{l|}{Line Represntation} & \multicolumn{1}{c|}{47} & 50 & \multicolumn{1}{c|}{10} & 18 \\
\multicolumn{1}{l|}{Stem Plot} & \multicolumn{1}{c|}{45} & 52 & \multicolumn{1}{c|}{6} & 15 \\
\multicolumn{1}{l|}{Stair Plot} & \multicolumn{1}{c|}{42} & 48 & \multicolumn{1}{c|}{10} & 26 \\
\multicolumn{1}{l|}{\begin{tabular}[c]{@{}l@{}}Ablation - removing \\ Y axis\end{tabular}} & \multicolumn{1}{c|}{63} & 64 & \multicolumn{1}{c|}{7} & 22 \\
\multicolumn{1}{l|}{Rotated X axis Tick} & \multicolumn{1}{c|}{56} & 62 & \multicolumn{1}{c|}{8} & 22 \\
\multicolumn{1}{l|}{Annotated Bar Graph} & \multicolumn{1}{c|}{77} & 82 & \multicolumn{1}{c|}{12} & 42 \\
\multicolumn{1}{l|}{Horizontal Bar Graph} & \multicolumn{1}{c|}{54} & 65 & \multicolumn{1}{c|}{11} & 19 \\ \bottomrule
\end{tabular}%
}
\caption{Effect of increasing font size}
\label{tab:font-tale}
\end{table}

\paragraph{Model Scores} Alongside the scores of models across various perturbations for complex charts provided in Table \ref{tab:pert-complex}, we have also presented the model scores for different perturbations in simple charts in Table \ref{tab:pert-simple}.

\paragraph{Where can models not answer?}
Continuing with the analysis of \textit{Q5} in section \ref{results}, we provide a further breakdown of cases where models failed to answer correctly, as shown in Table \ref{errorRate}. In addition to previously discussed issues, we found that in some instances (4/181), models did not fully comprehend the entire chart before answering. 

\begin{table}[!ht]
\begin{center}
    
\resizebox{0.9\columnwidth}{!}{%
\begin{tabular}{c|c|c|c}
\hline
Type of chart & Wrong & Total & Error Rate  \\ \hline
Bar           & 121   & 1842  & 6.56\%      \\
Line          & 44    & 380   & \textbf{11.57}\% \\
Pie           & 16    & 151   & 10.59\%     \\ \hline
Total         & 181   & 2373  & 7.62\%      \\ \hline
\end{tabular}%
}
\end{center}
\caption{Error distribution among different chart types}
\label{errorRate}
\end{table}

For example, in a chart with four columns labeled “No (low confidence), No (high confidence), Yes (low confidence), and Yes (high confidence),” models were asked to calculate the percentage of 'No'. However, they failed to recognize that the question required summing the percentages from both ‘No’ columns and instead provided the percentage from only one column. This error highlights a potential gap in the models' ability to fully integrate image encoding with language decoding, suggesting improvements could be made to better interpret such visual data.

\begin{table*}[!htbp]
\resizebox{\textwidth}{!}{%
\begin{tabular}{ccccc}
\hline
\multicolumn{5}{c}{\textit{\textbf{Models and Perturbation Types}}}                                                                                                                                                                                                                                      \\ \hline
\multicolumn{1}{c|}{\textbf{Gemini 1.5 Flash}}                                                    & \multicolumn{1}{c|}{\textbf{GPT-4o}}                                                              & \multicolumn{1}{c|}{\textbf{Qwen-VL}}                                     & \multicolumn{1}{c|}{\textbf{CogAgent-VQA}}                                                       & \textbf{InternLM-XComposer2}                                                \\ \hline
\multicolumn{1}{c|}{\begin{tabular}[c]{@{}c@{}}Annotations on \\ individual points\end{tabular}}  & \multicolumn{1}{c|}{\begin{tabular}[c]{@{}c@{}}Annotations on \\ individual points\end{tabular}}  & \multicolumn{1}{c|}{\begin{tabular}[c]{@{}c@{}}Annotations on Bar \\ Graphs\end{tabular}}         & \multicolumn{1}{c|}{\begin{tabular}[c]{@{}c@{}}Annotations on \\ individual points\end{tabular}} & \begin{tabular}[c]{@{}c@{}}Annotations on bar \\ charts\end{tabular}        \\
\multicolumn{1}{c|}{\begin{tabular}[c]{@{}c@{}}Annotations on Bar \\ Graphs\end{tabular}}         & \multicolumn{1}{c|}{\begin{tabular}[c]{@{}c@{}}Annotations on Bar \\ Graphs\end{tabular}}         & \multicolumn{1}{c|}{\begin{tabular}[c]{@{}c@{}}Annotations on \\ individual points\end{tabular}}  & \multicolumn{1}{c|}{\begin{tabular}[c]{@{}c@{}}Annotations on Bar \\ Graphs\end{tabular}}        & \begin{tabular}[c]{@{}c@{}}Annotations on \\ individual points\end{tabular} \\
\multicolumn{1}{c|}{\begin{tabular}[c]{@{}c@{}}Random Color Scheme\\ in Chart\end{tabular}}       & \multicolumn{1}{c|}{\begin{tabular}[c]{@{}c@{}}Placing Legend \\ Elements with Line\end{tabular}} & \multicolumn{1}{c|}{\begin{tabular}[c]{@{}c@{}}Basic Matplotlib \\ Charts\end{tabular}}           & \multicolumn{1}{c|}{\begin{tabular}[c]{@{}c@{}}Random Color Scheme\\ in Chart\end{tabular}}      & Area Plot                                                                   \\
\multicolumn{1}{c|}{\begin{tabular}[c]{@{}c@{}}Placing Legend \\ Elements with Line\end{tabular}} & \multicolumn{1}{c|}{\begin{tabular}[c]{@{}c@{}}Random Markers \\ and Line Styles\end{tabular}}    & \multicolumn{1}{c|}{\begin{tabular}[c]{@{}c@{}}Placing Legend \\ Elements with Line\end{tabular}} & \multicolumn{1}{c|}{Axes Transposition}                                                          & \begin{tabular}[c]{@{}c@{}}Horizontal Bar \\ Charts\end{tabular}            \\
\multicolumn{1}{c|}{\begin{tabular}[c]{@{}c@{}}Basic Matplotlib \\ Charts\end{tabular}}           & \multicolumn{1}{c|}{\begin{tabular}[c]{@{}c@{}}Basic Matplotlib \\ Charts\end{tabular}}           & \multicolumn{1}{c|}{Changing Font Size}                                                           & \multicolumn{1}{c|}{\begin{tabular}[c]{@{}c@{}}Basic Matplotlib \\ Charts\end{tabular}}          & Random Color Scheme                                                         \\ \hline
\end{tabular}%
}
\caption{Top 5 best performing perturbations for each model}
\label{tab:best5}
\end{table*}

\begin{table*}[!htbp]
\resizebox{\textwidth}{!}{%
\begin{tabular}{ccccc}
\hline
\multicolumn{5}{c}{\textit{\textbf{Models and Perturbation Types}}}                                                                                                                                                                                                                                                                                                                                                                                                                                               \\ \hline
\multicolumn{1}{c|}{\textbf{Gemini 1.5 Flash}}                                           & \multicolumn{1}{c|}{\textbf{GPT-4o}}                                                          & \multicolumn{1}{c|}{\textbf{Qwen-VL}}                                                                     & \multicolumn{1}{c|}{\textbf{CogAgent-VQA}}                                                    & \multicolumn{1}{c}{\textbf{InternLM-Xcomposer2}}                                                         \\ \hline
\multicolumn{1}{c|}{Stacked Area Chart}                                                  & \multicolumn{1}{c|}{\begin{tabular}[c]{@{}c@{}}Horizontally Stacked\\ Bars\end{tabular}}      & \multicolumn{1}{c|}{Stacked Bar Graphs}                                                                   & \multicolumn{1}{c|}{\begin{tabular}[c]{@{}c@{}}Horizontal Bar \\ Charts\end{tabular}}         & \multicolumn{1}{c}{\begin{tabular}[c]{@{}c@{}}Horizontally Stacked\\ Bars\end{tabular}}                  \\
\multicolumn{1}{c|}{\begin{tabular}[c]{@{}c@{}}Horizontally Stacked\\ Bars\end{tabular}} & \multicolumn{1}{c|}{Stacked Area Chart}                                                       & \multicolumn{1}{c|}{\begin{tabular}[c]{@{}c@{}}Changing Horizontal\\ and Vertical Dimension\end{tabular}} & \multicolumn{1}{c|}{Stacked Area Chart}                                                       & \multicolumn{1}{c}{\begin{tabular}[c]{@{}c@{}}Changing Horizontal\\ and Vertical Dimension\end{tabular}} \\
\multicolumn{1}{c|}{Stacked Bar Graphs}                                                  & \multicolumn{1}{c|}{Stacked Bar Graphs}                                                       & \multicolumn{1}{c|}{Log Scale}                                                                            & \multicolumn{1}{c|}{\begin{tabular}[c]{@{}c@{}}Horizontally Stacked\\ Bars\end{tabular}}      & \multicolumn{1}{c}{Stacked Area Chart}                                                                   \\
\multicolumn{1}{c|}{Log Scale}                                                           & \multicolumn{1}{c|}{\begin{tabular}[c]{@{}c@{}}Horizontal Grouped \\ Bar Charts\end{tabular}} & \multicolumn{1}{c|}{\begin{tabular}[c]{@{}c@{}}Random Representation\\ of Scatter Plots\end{tabular}}     & \multicolumn{1}{c|}{\begin{tabular}[c]{@{}c@{}}Horizontal Grouped \\ Bar Charts\end{tabular}} & \multicolumn{1}{c}{Stacked Bar Graphs}                                                                   \\
\multicolumn{1}{c|}{Normal Stair Plot}                                                   & \multicolumn{1}{c|}{\begin{tabular}[c]{@{}c@{}}Hatched Pattern in\\ Bar Charts\end{tabular}}  & \multicolumn{1}{c|}{Stair Plots with Marker}                                                              & \multicolumn{1}{c|}{Log Scale}                                                                & \multicolumn{1}{c}{Changing Font Size}                                                                   \\ \hline
\end{tabular}%
}
\caption{Top 5 worst performing perturbations for each model}
\label{tab:worst5}
\end{table*}

\begin{table*}[!htbp]
\small
\setlength{\tabcolsep}{1.0pt}
\begin{tabular}{@{}cccccc|ccccc@{}}
\toprule
 & \multicolumn{5}{c|}{Simple Questions} & \multicolumn{5}{c}{Complex Questions} \\ \midrule
 & \multicolumn{2}{|c|}{MLLMs} & \multicolumn{3}{c|}{Generalist VLMs} & \multicolumn{2}{c|}{MLLMs} & \multicolumn{3}{c}{Generalist VLMs} \\ \midrule
 \multicolumn{1}{c|}{Category}                        & \begin{tabular}[c]{@{}c@{}}Gemini \\ 1.5 Flash\end{tabular} & \multicolumn{1}{c|}{GPT-4o}      & \begin{tabular}[c]{@{}c@{}} Qwen \\ VL \end{tabular}    & \begin{tabular}[c]{@{}c@{}}CogAgent\\ VQA\end{tabular} & \begin{tabular}[c]{@{}c@{}}InternLM\\ XComposer2\end{tabular} & \begin{tabular}[c]{@{}c@{}}Gemini \\ 1.5 Flash\end{tabular} & \multicolumn{1}{c|}{GPT-4o}               & \begin{tabular}[c]{@{}c@{}} Qwen \\ VL \end{tabular}    & \begin{tabular}[c]{@{}c@{}}CogAgent\\ VQA\end{tabular} & \begin{tabular}[c]{@{}c@{}}InternLM\\ XComposer2\end{tabular} \\ 
 \midrule
 
\multicolumn{1}{c|}{\textit{original\_chart}} & \textit{\textbf{96}} & \multicolumn{1}{c|}{\textit{94}} & \textit{76} & \textit{79} & \textit{83} & \textit{85} & \multicolumn{1}{c|}{\textit{\textbf{89}}} & \textit{56} & \textit{64} & \textit{69} \\ \midrule
\multicolumn{1}{c|}{annotations} & 90 & \multicolumn{1}{c|}{\textbf{91}} & 62 & 66 & 65 & \textbf{74} & \multicolumn{1}{c|}{64} & 42 & 42 & 47 \\
\multicolumn{1}{c|}{area\_plot} & \textbf{73} & \multicolumn{1}{c|}{42} & 21 & 16 & 61 & \textbf{71} & \multicolumn{1}{c|}{64} & 39 & 42 & 48 \\
\multicolumn{1}{c|}{annotated\_bars} & \textbf{93} & \multicolumn{1}{c|}{91} & 71 & 63 & 73 & 78 & \multicolumn{1}{c|}{\textbf{90}} & 45 & 58 & 48 \\
\multicolumn{1}{c|}{basic} & \textbf{73} & \multicolumn{1}{c|}{43} & 24 & 18 & 54 & \textbf{73} & \multicolumn{1}{c|}{61} & 38 & 38 & 46 \\
\multicolumn{1}{c|}{color\_random} & \textbf{79} & \multicolumn{1}{c|}{43} & 20 & 22 & 63 & \textbf{72} & \multicolumn{1}{c|}{64} & 32 & 40 & 42 \\
\multicolumn{1}{c|}{color\_scheme} & \textbf{78} & \multicolumn{1}{c|}{50} & 18 & 23 & 58 & \textbf{68} & \multicolumn{1}{c|}{64} & 30 & 36 & 36 \\
\multicolumn{1}{c|}{data\_pivot} & \textbf{74} & \multicolumn{1}{c|}{55} & 13 & 13 & 56 & \textbf{71} & \multicolumn{1}{c|}{62} & 42 & 40 & 41 \\
\multicolumn{1}{c|}{font} & \textbf{79} & \multicolumn{1}{c|}{53} & 19 & 28 & 28 & \textbf{65} & \multicolumn{1}{c|}{52} & 33 & 26 & 51 \\
\multicolumn{1}{c|}{grid} & \textbf{79} & \multicolumn{1}{c|}{58} & 23 & 24 & 57 & \textbf{66} & \multicolumn{1}{c|}{64} & 31 & 42 & 44 \\
\multicolumn{1}{c|}{hatching} & \textbf{75} & \multicolumn{1}{c|}{44} & 20 & 18 & 67 & \textbf{72} & \multicolumn{1}{c|}{69} & 39 & 42 & 44 \\
\multicolumn{1}{c|}{horizontal} & \textbf{73} & \multicolumn{1}{c|}{33} & 19 & 14 & 58 & \textbf{67} & \multicolumn{1}{c|}{64} & 35 & 45 & 43 \\
\multicolumn{1}{c|}{legend\_position} & \textbf{78} & \multicolumn{1}{c|}{57} & 14 & 23 & 54 & 59 & \multicolumn{1}{c|}{\textbf{62}} & 32 & 41 & 43 \\
\multicolumn{1}{c|}{line\_representation} & \textbf{84} & \multicolumn{1}{c|}{59} & 19 & 25 & 56 & \textbf{67} & \multicolumn{1}{c|}{63} & 31 & 34 & 40 \\
\multicolumn{1}{c|}{log\_scale} & \textbf{42} & \multicolumn{1}{c|}{36} & 12 & 12 & 14 & \textbf{78} & \multicolumn{1}{c|}{21} & 32 & 37 & 45 \\
\multicolumn{1}{c|}{replacing\_legend\_with\_labels} & \textbf{78} & \multicolumn{1}{c|}{59} & 18 & 22 & 53 & \textbf{72} & \multicolumn{1}{c|}{61} & 27 & 41 & 40 \\
\multicolumn{1}{c|}{scaling\_size} & \textbf{73} & \multicolumn{1}{c|}{53} & 17 & 25 & 31 & \textbf{62} & \multicolumn{1}{c|}{55} & 34 & 38 & 39 \\
\multicolumn{1}{c|}{scatter\_representation} & \textbf{75} & \multicolumn{1}{c|}{48} & 15 & 17 & 47 & \textbf{64} & \multicolumn{1}{c|}{57} & 34 & 43 & 44 \\
\multicolumn{1}{c|}{stair\_plot\_normal} & \textbf{59} & \multicolumn{1}{c|}{53} & 20 & 23 & 52 & \textbf{65} & \multicolumn{1}{c|}{61} & 39 & 38 & 48 \\
\multicolumn{1}{c|}{stair\_plot\_with\_marker} & \textbf{64} & \multicolumn{1}{c|}{51} & 15 & 24 & 60 & \textbf{68} & \multicolumn{1}{c|}{64} & 25 & 41 & 31 \\
\multicolumn{1}{c|}{stem\_plot} & \textbf{72} & \multicolumn{1}{c|}{41} & 12 & 17 & 70 & 75 & \multicolumn{1}{c|}{\textbf{86}} & 36 & 54 & 48 \\
\multicolumn{1}{c|}{tick\_orientation} & \textbf{76} & \multicolumn{1}{c|}{57} & 18 & 22 & 50 & \textbf{72} & \multicolumn{1}{c|}{60} & 41 & 46 & 49 \\
\multicolumn{1}{c|}{tick\_position} & \textbf{69} & \multicolumn{1}{c|}{54} & 27 & 27 & 51 & 53 & \multicolumn{1}{c|}{\textbf{61}} & 35 & 42 & 41 \\ \bottomrule
\end{tabular}%
\caption{Model Performance on various perturbations on Simple Charts.}
\label{tab:pert-simple}
\end{table*}

\clearpage

\begin{figure*}
    \centering
    \includegraphics[width = \linewidth]{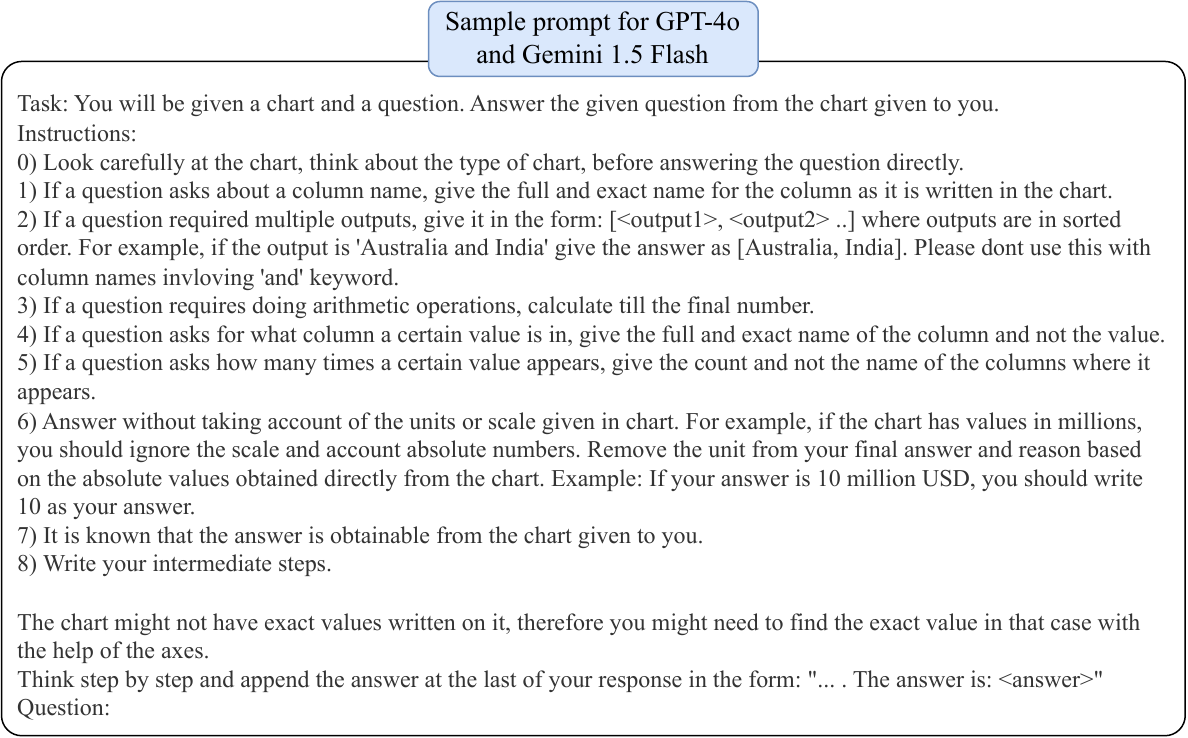}
    \caption{Prompt for testing chart question answering}
    \label{fig:example_prompt_qa}
\end{figure*}

\begin{figure*}
    \centering
    \includegraphics[width = \linewidth]{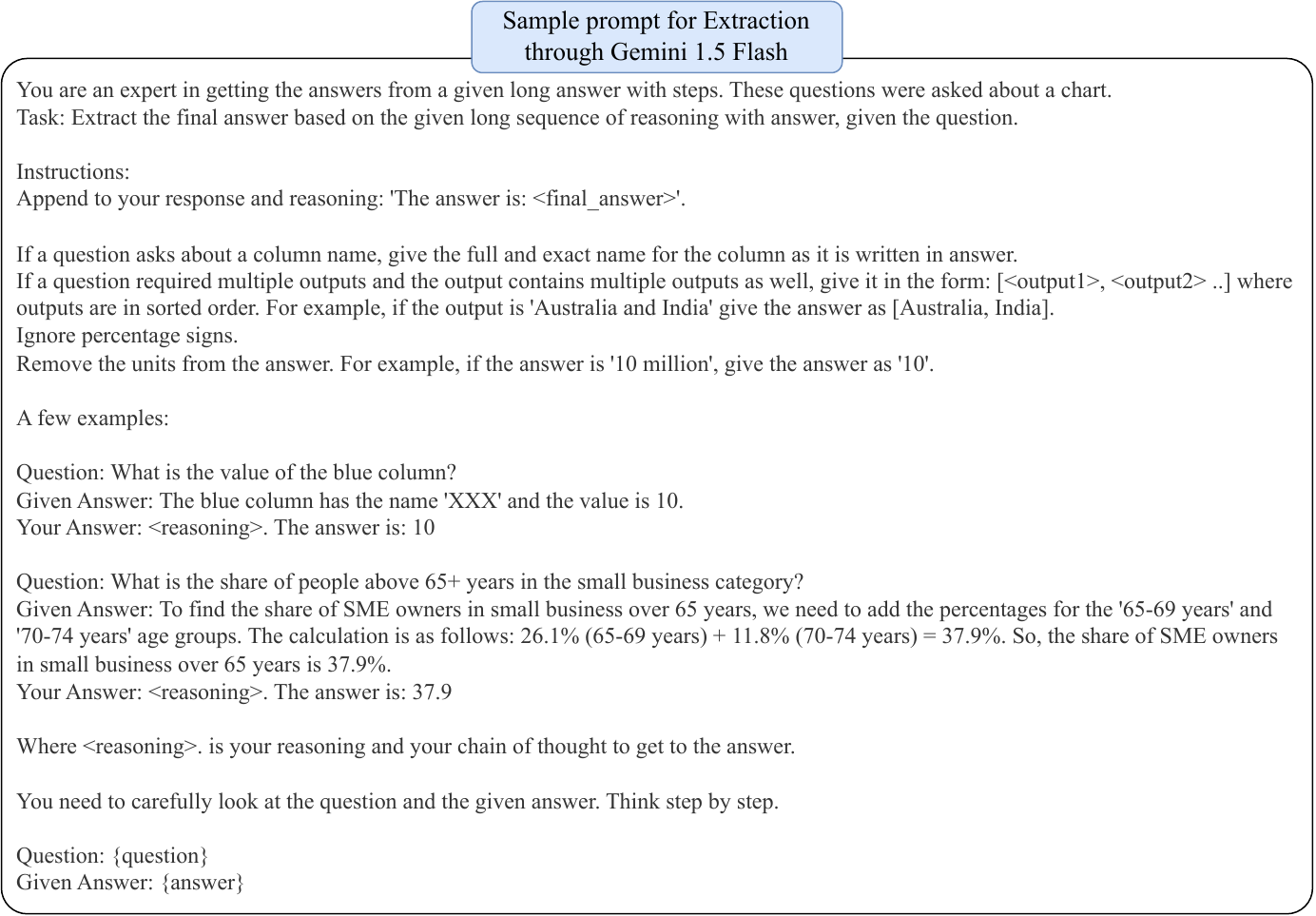}
    \caption{Prompt for extracting answers through an LLM from a different LLM}
    \label{fig:extractorPrompt}    
\end{figure*}

\begin{figure*}
    \centering
    \includegraphics[width = 0.6\linewidth]{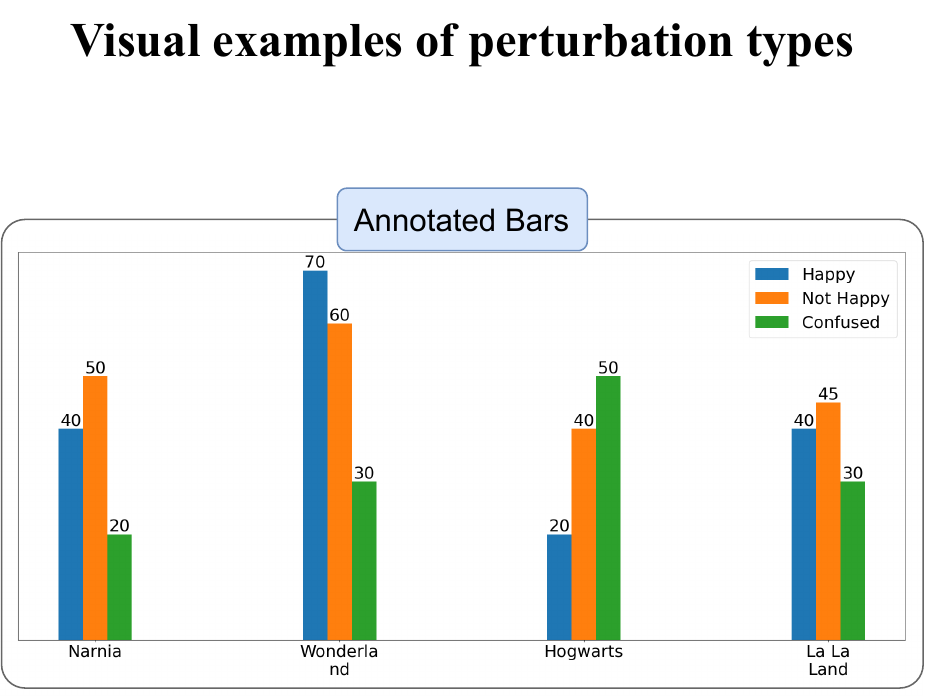}
    \label{fig:annotated_bars}
\end{figure*}

\begin{figure*}
    \centering
    \includegraphics[width = \linewidth]{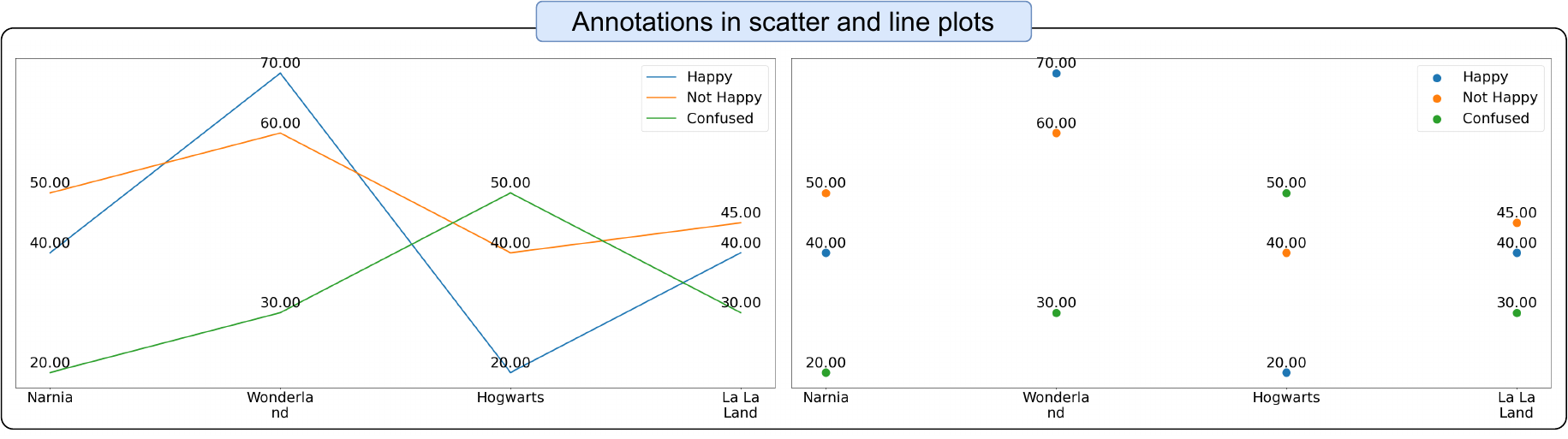}
    \label{fig:annotations}
\end{figure*}

\begin{figure*}
    \centering
    \includegraphics[width = \linewidth]{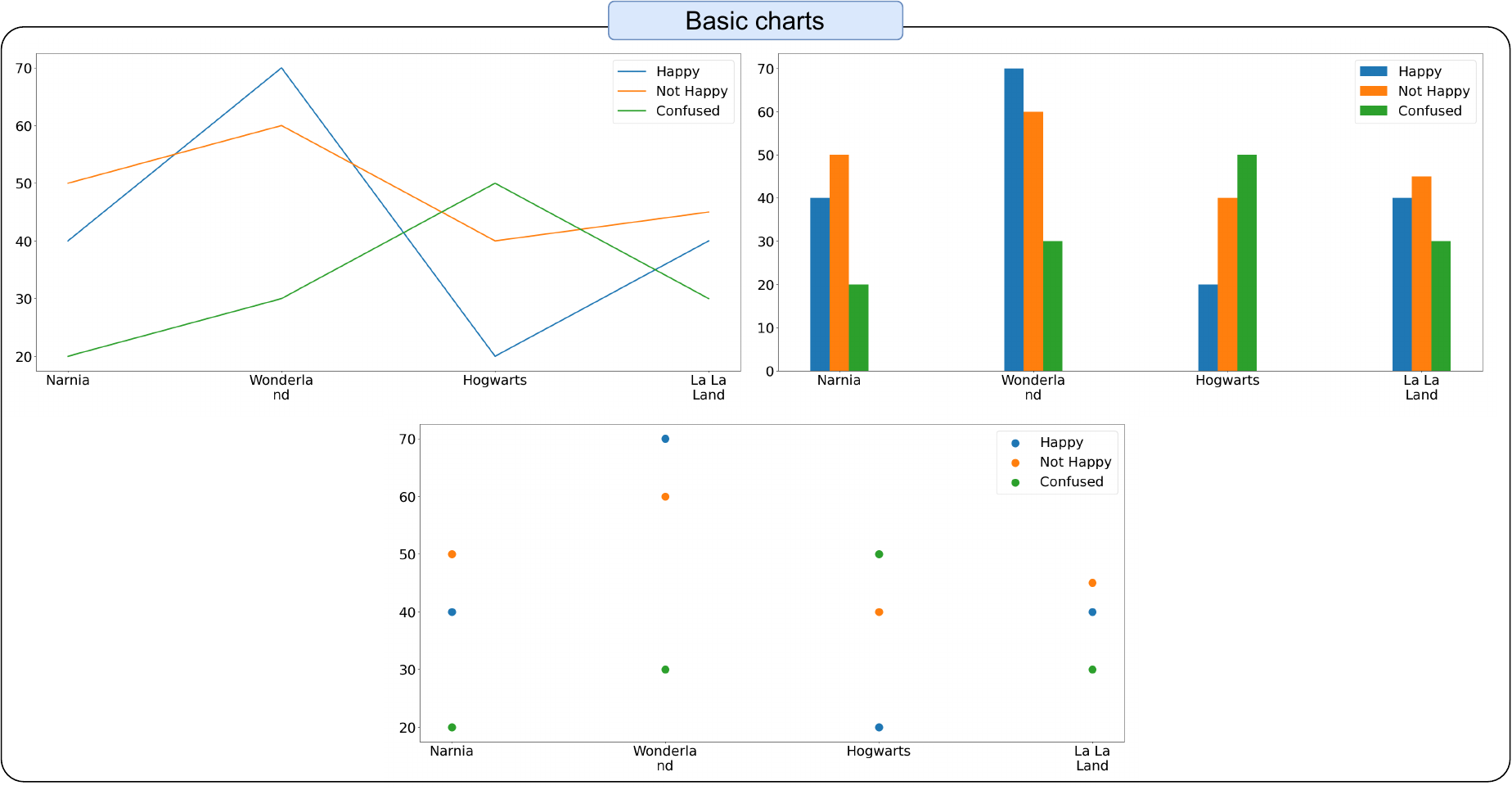}
    \label{fig:basic}
\end{figure*}

\begin{figure*}
    \centering
    \includegraphics[width = \linewidth]{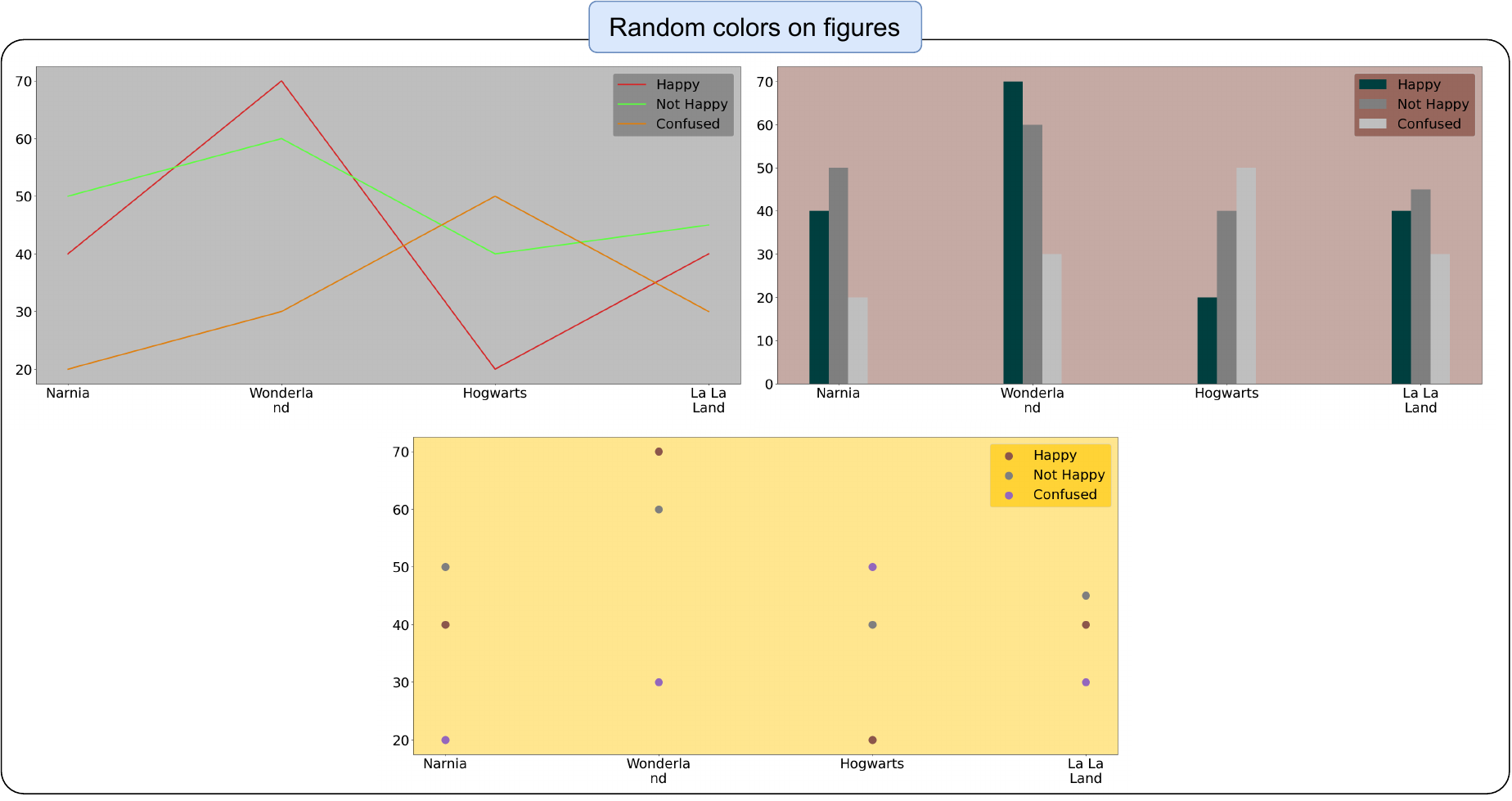}
    \label{fig:color_random}
\end{figure*}

\begin{figure*}
    \centering
    \includegraphics[width = \linewidth]{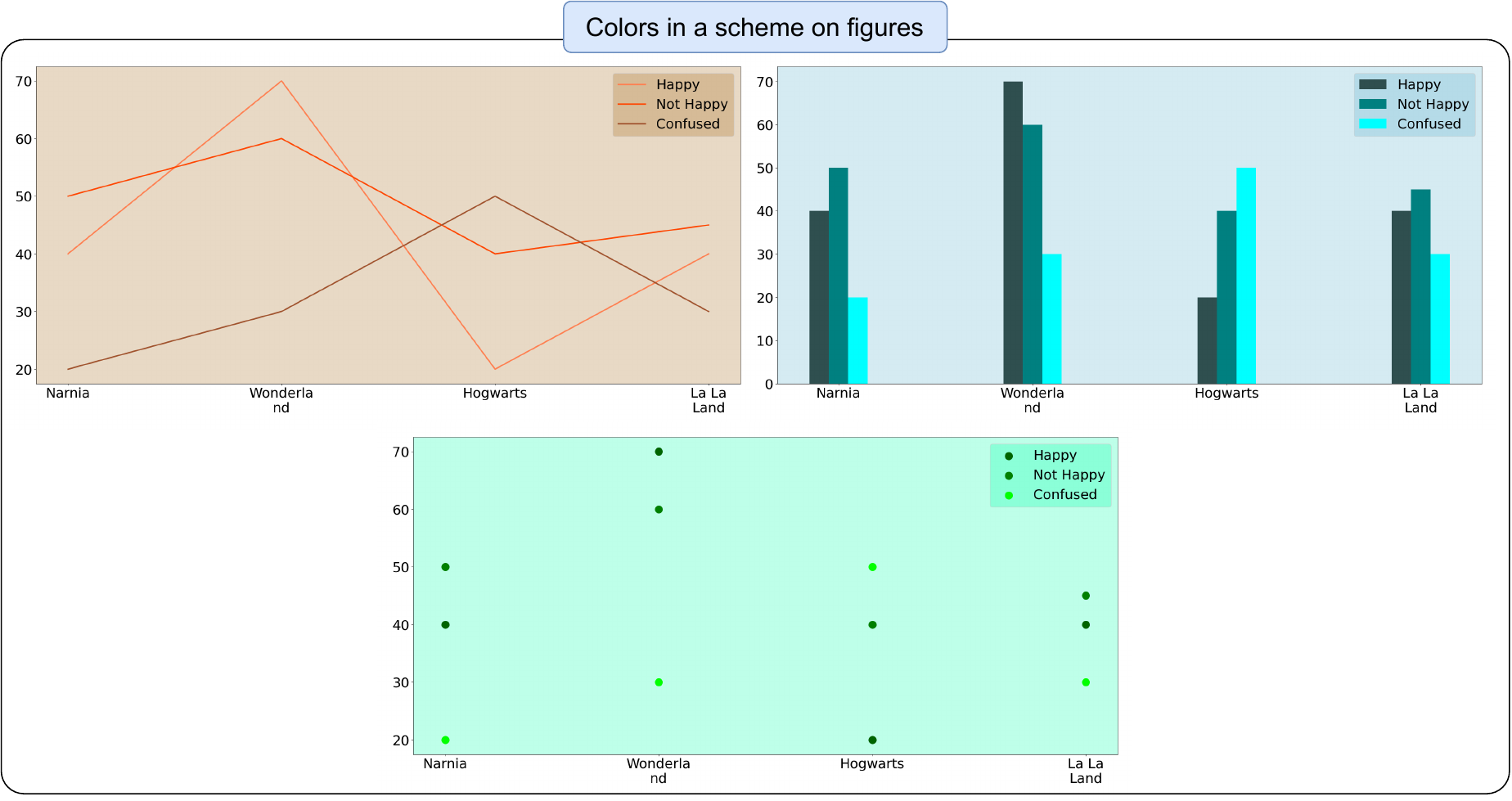}
    \label{fig:cs}
\end{figure*}

\begin{figure*}
    \centering
    \includegraphics[width = \linewidth]{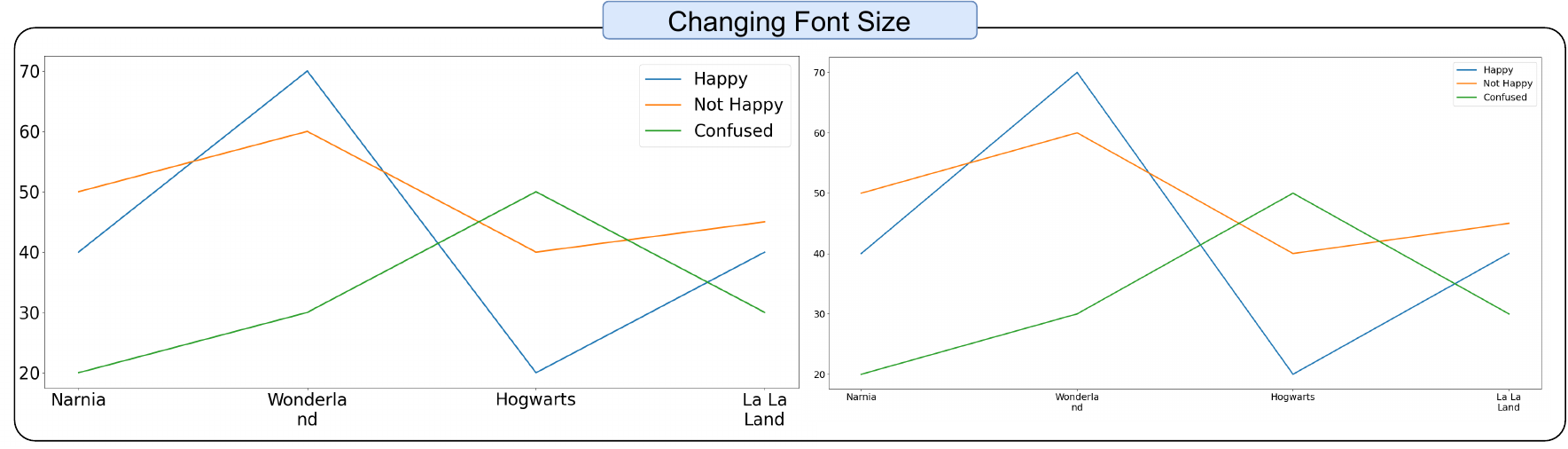}
    \label{fig:font}
\end{figure*}

\begin{figure*}
    \centering
    \includegraphics[width = \linewidth]{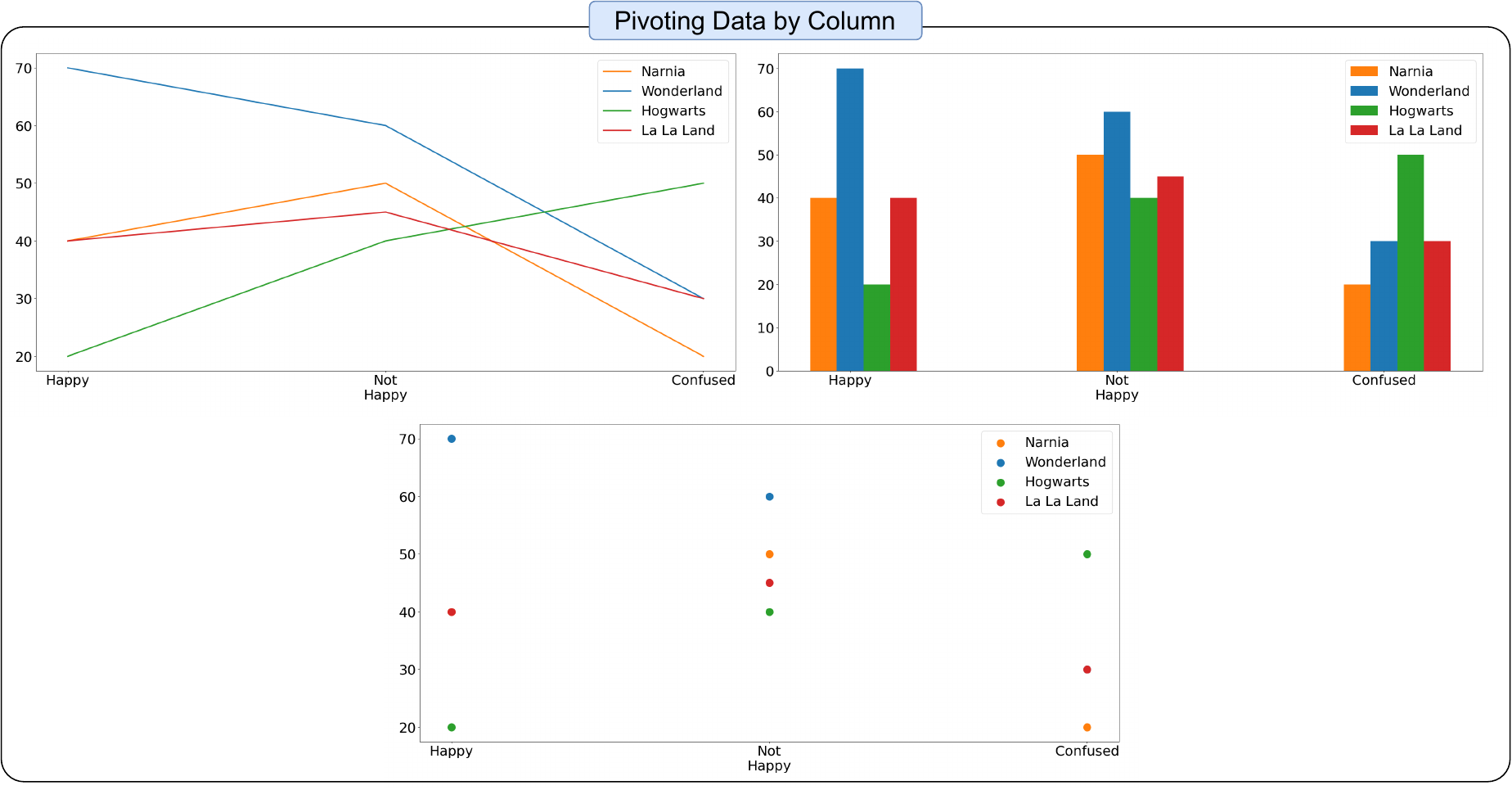}
    \label{fig:data_piv}
\end{figure*}

\begin{figure*}
    \centering
    \includegraphics[width = \linewidth]{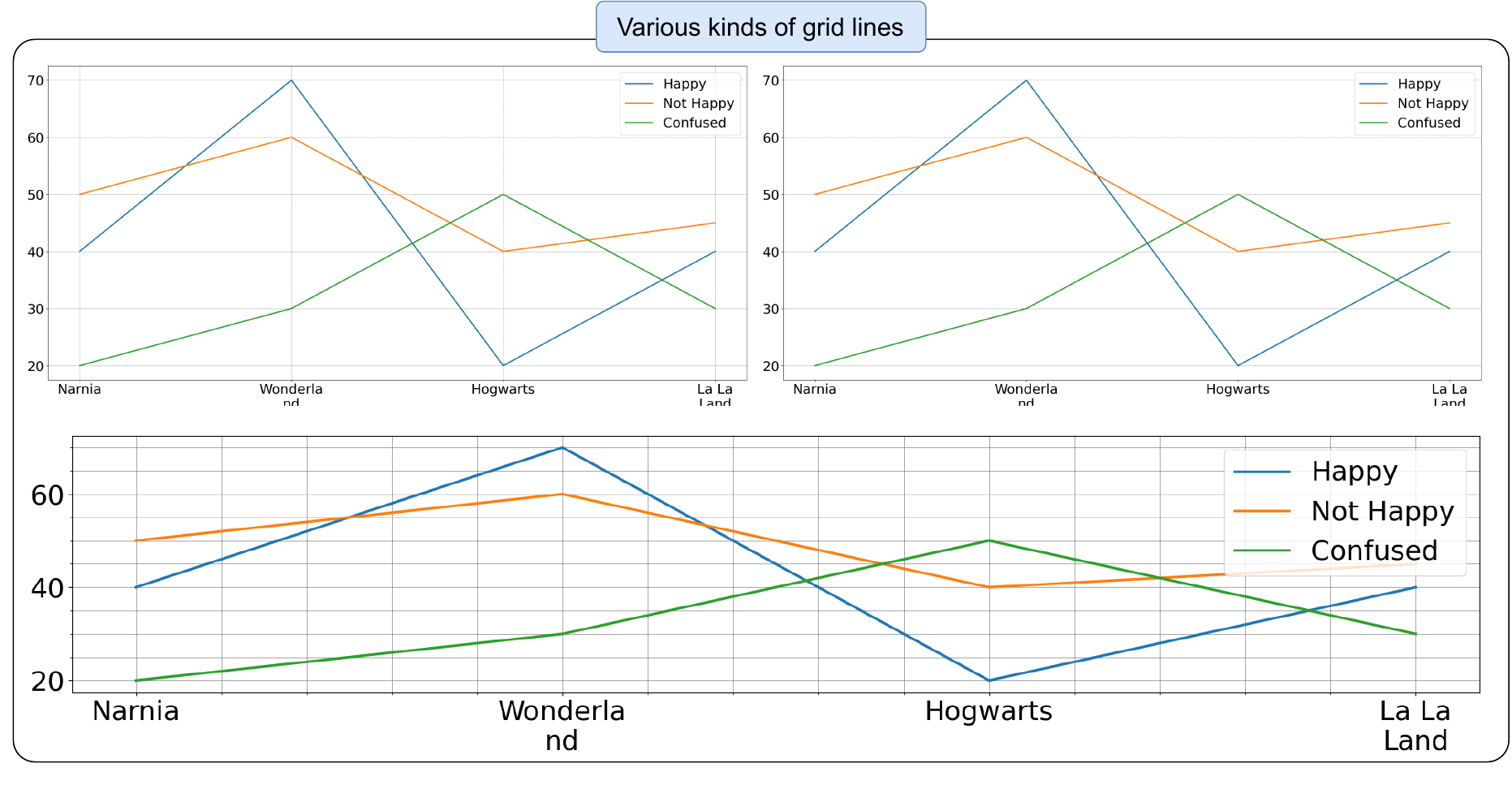}
    \label{fig:grid}
\end{figure*}

\begin{figure*}
    \centering
    \includegraphics[width = \linewidth]{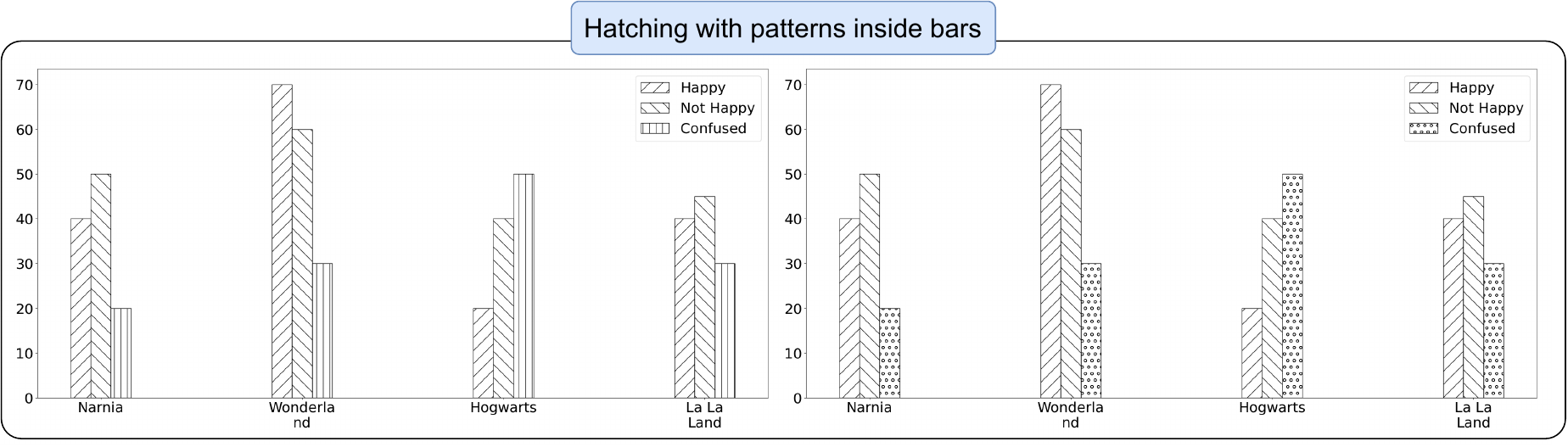}
    \label{fig:hatch}
\end{figure*}

\begin{figure*}
    \centering
    \includegraphics[width = \linewidth]{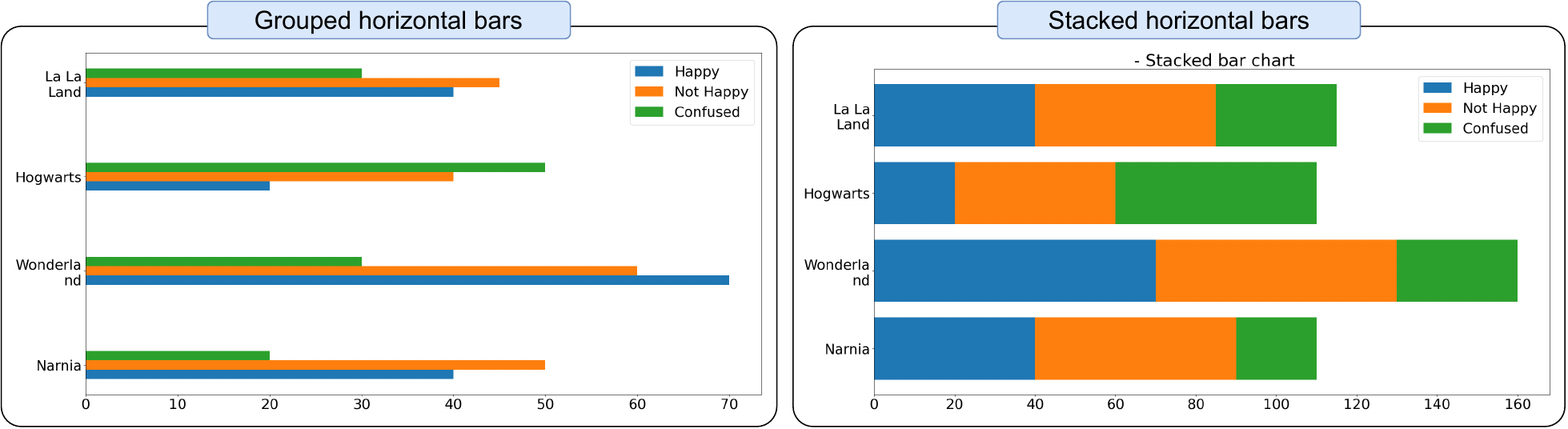}
    \label{fig:hg}
\end{figure*}

\begin{figure*}
    \centering
    \includegraphics[width = \linewidth]{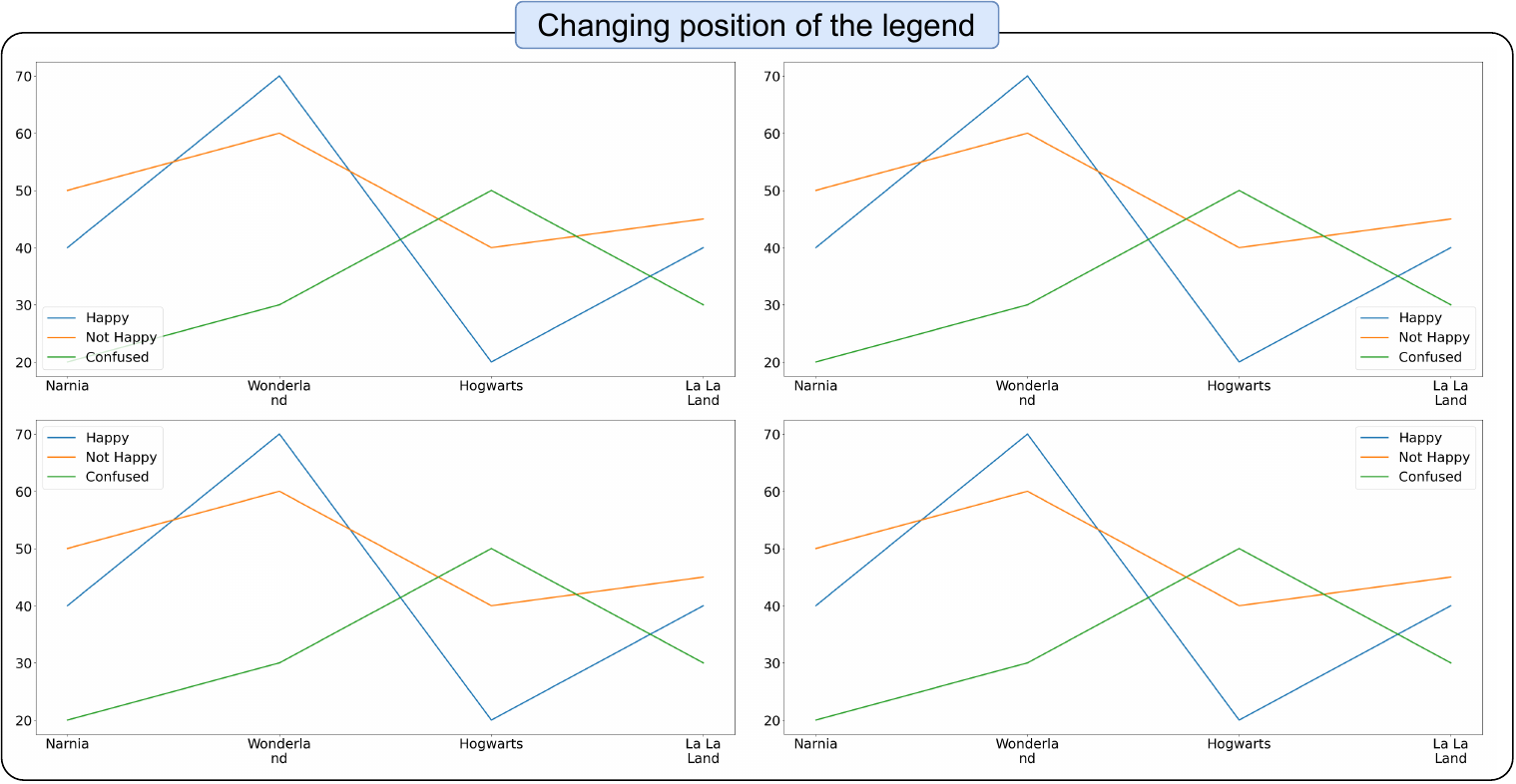}
    \label{fig:lp}
\end{figure*}

\begin{figure*}
    \centering
    \includegraphics[width = \linewidth]{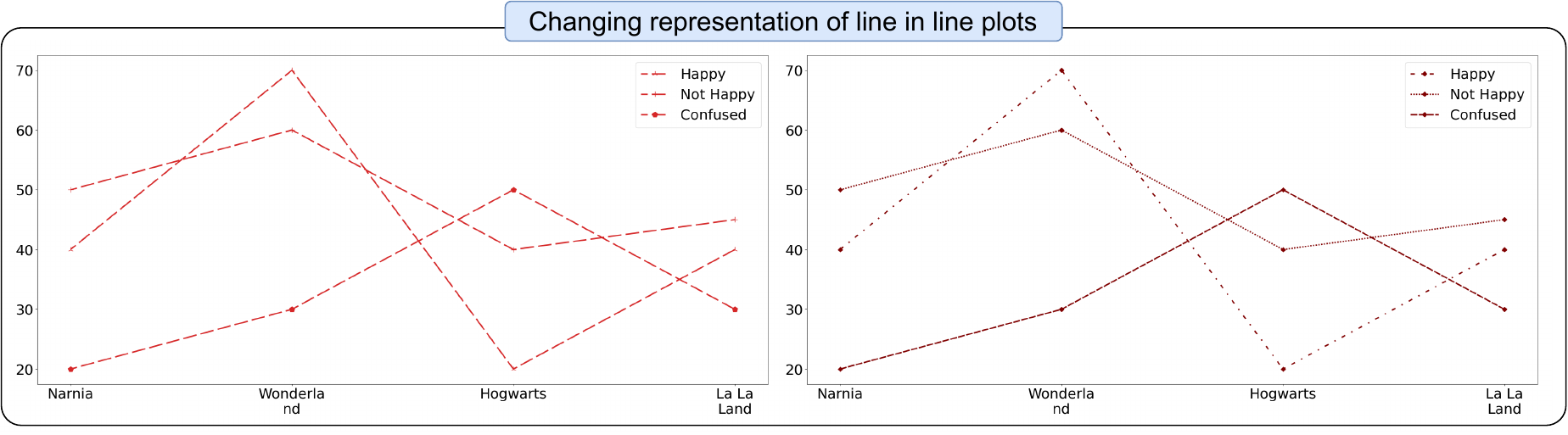}
    \label{fig:lr}
\end{figure*}

\begin{figure*}
    \centering
    \includegraphics[width = \linewidth]{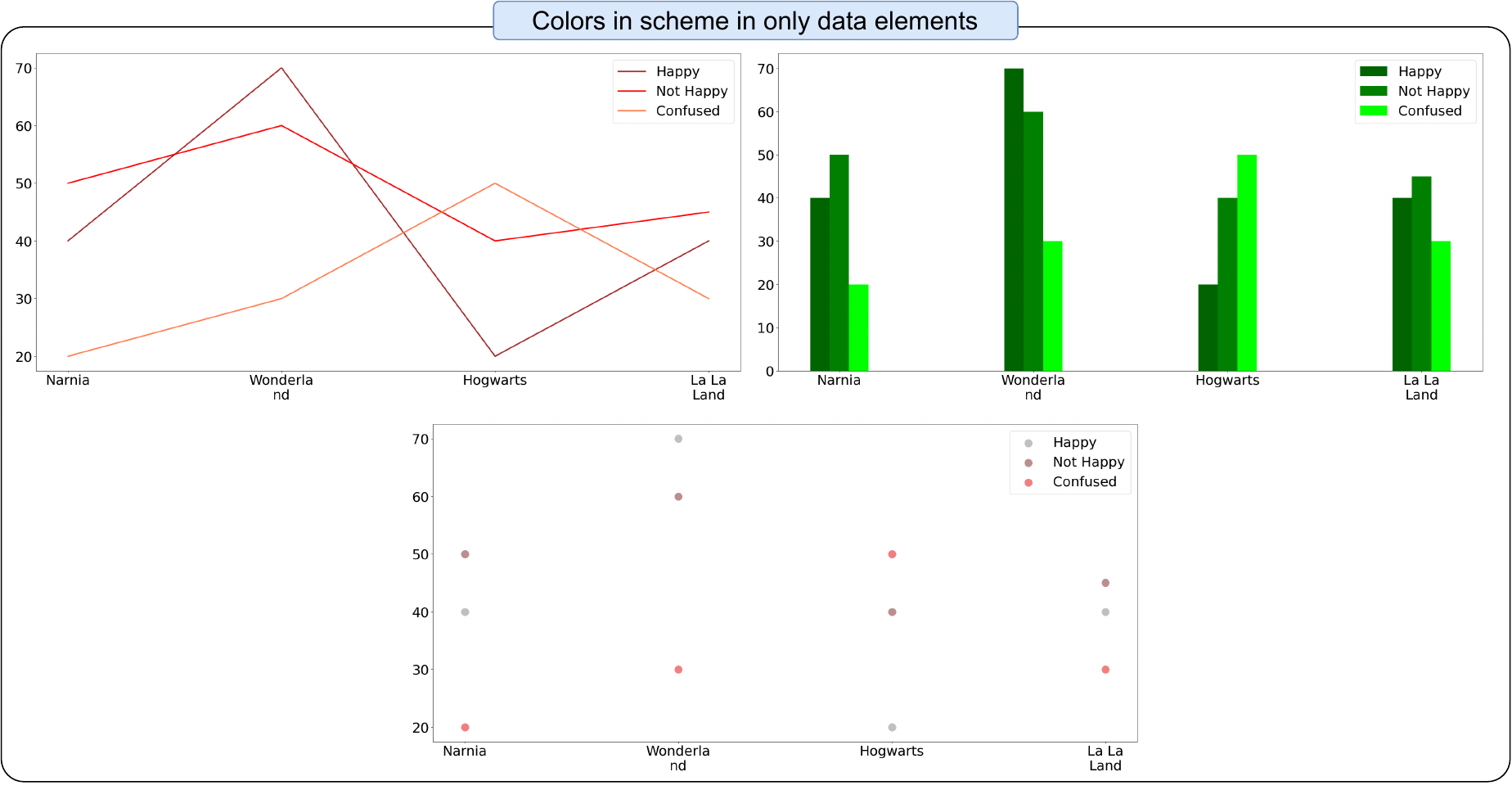}
    \label{fig:odcs}
\end{figure*}

\begin{figure*}
    \centering
    \includegraphics[width = \linewidth]{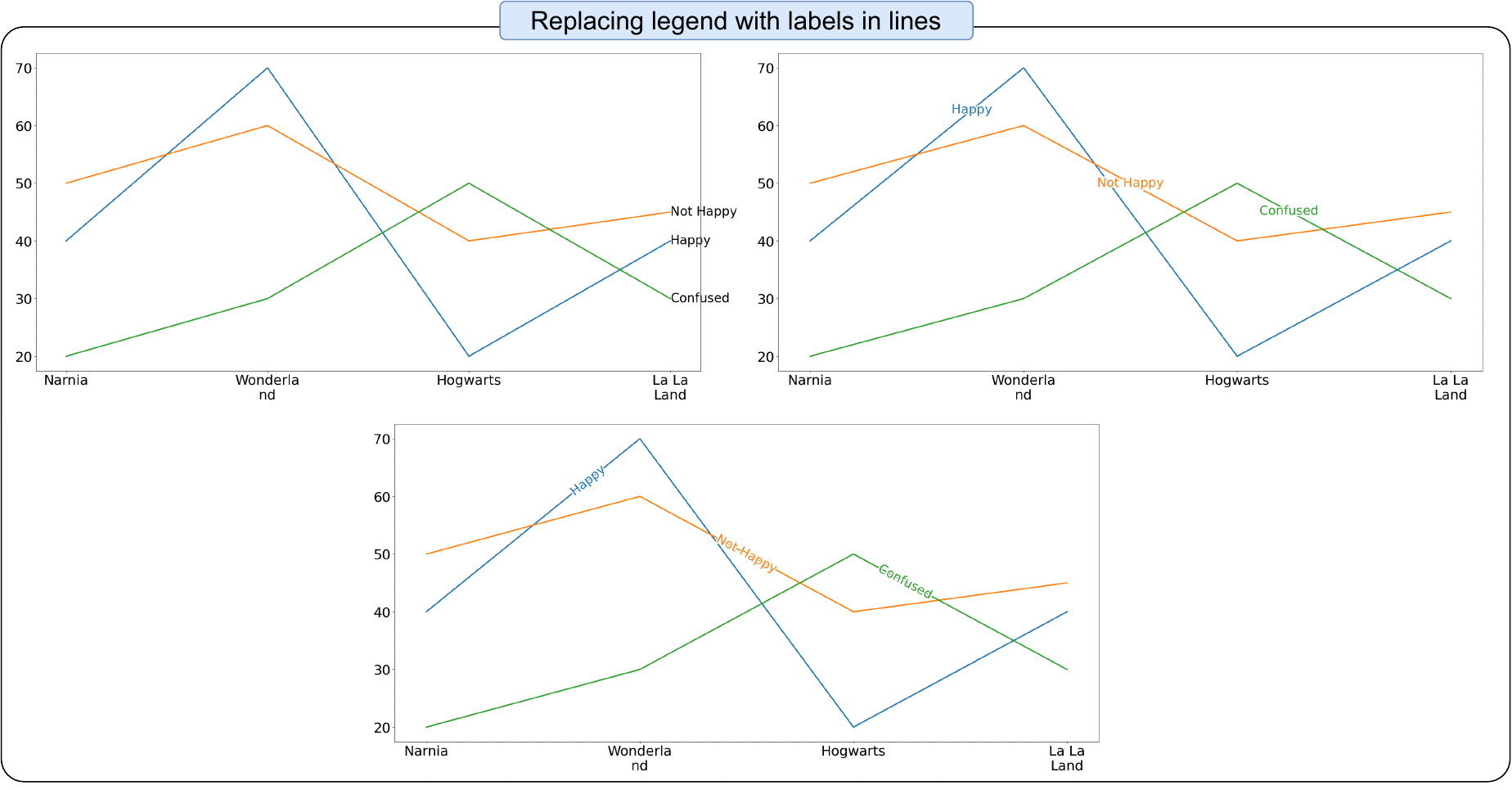}
    \label{fig:leg_rem}
\end{figure*}

\begin{figure*}
    \centering
    \includegraphics[width = \linewidth]{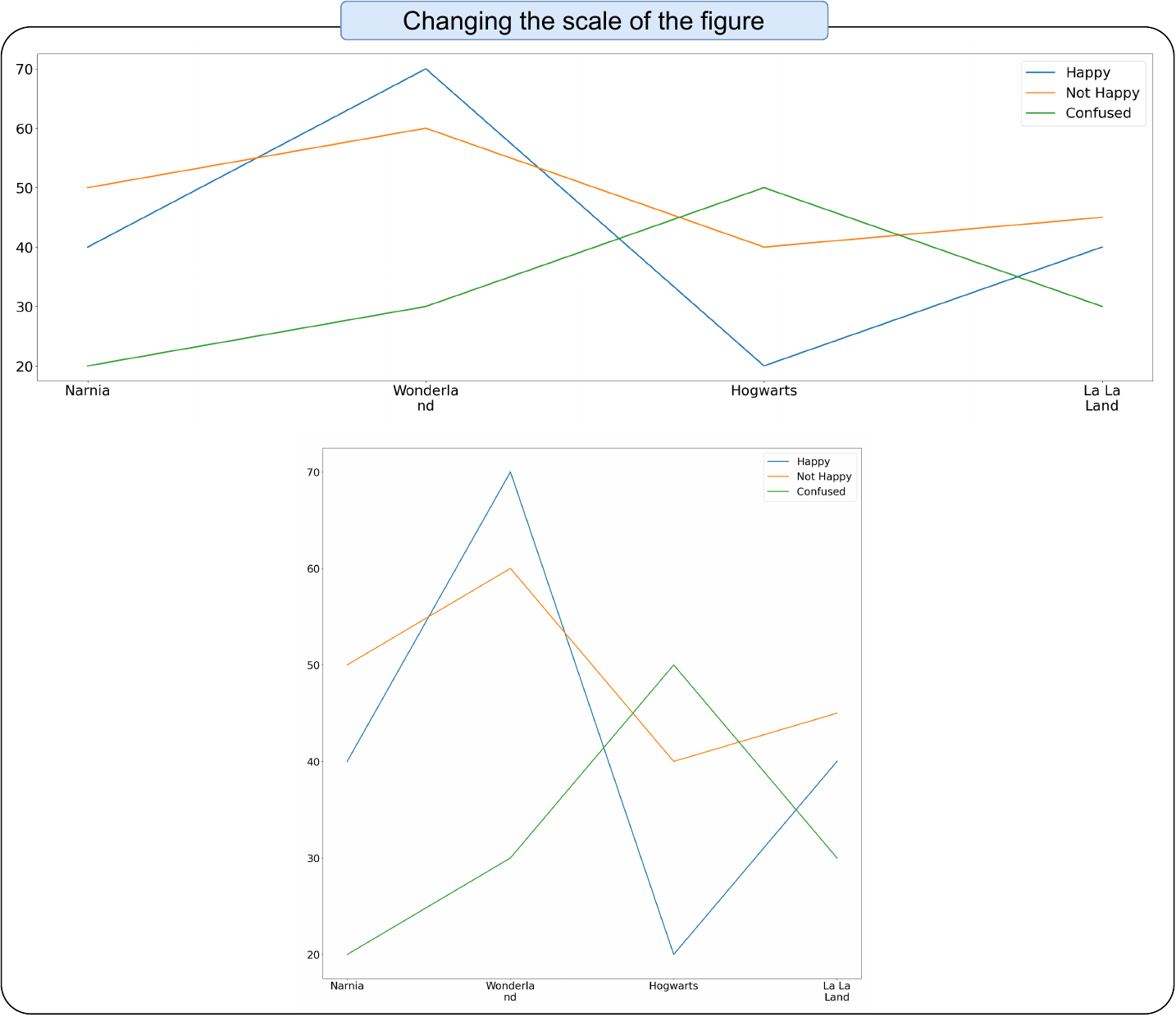}
    \label{fig:scaling}
\end{figure*}

\begin{figure*}
    \centering
    \includegraphics[width = \linewidth]{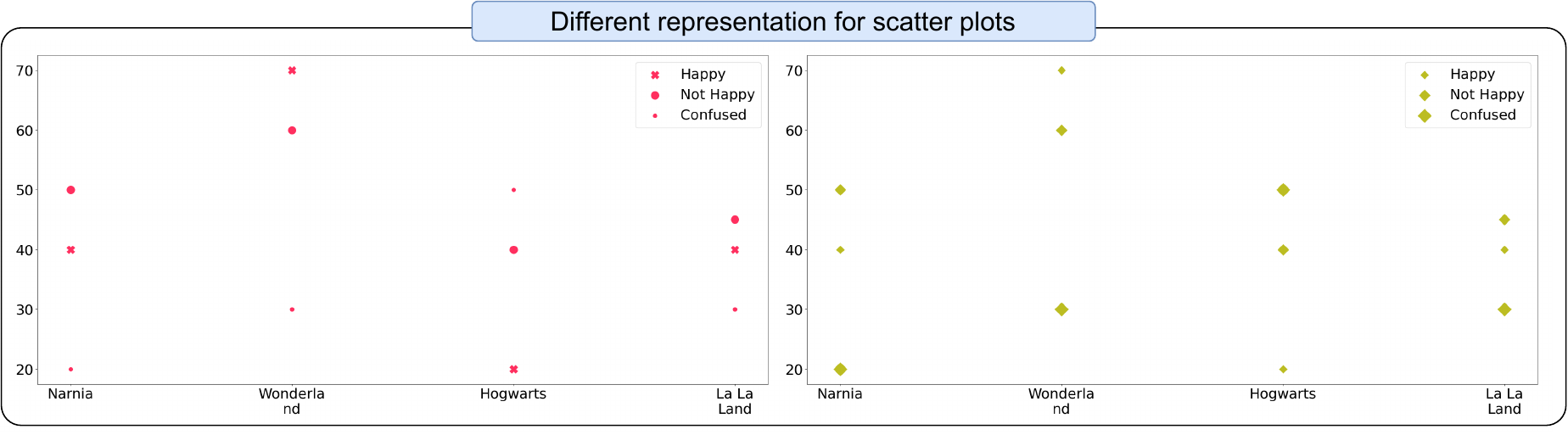}
    \label{fig:scatt}
\end{figure*}

\begin{figure*}
    \centering
    \includegraphics[width = \linewidth]{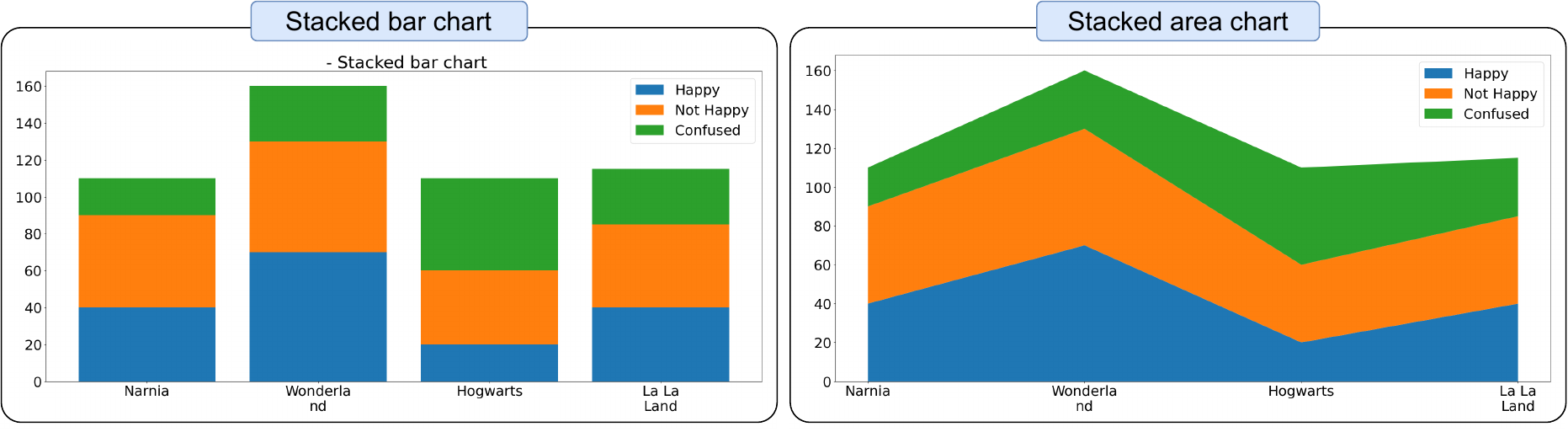}
    \caption{Stacked: Stacked bar graphs}
    \label{fig:stack}
\end{figure*}

\begin{figure*}
    \centering
    \includegraphics[width = \linewidth]{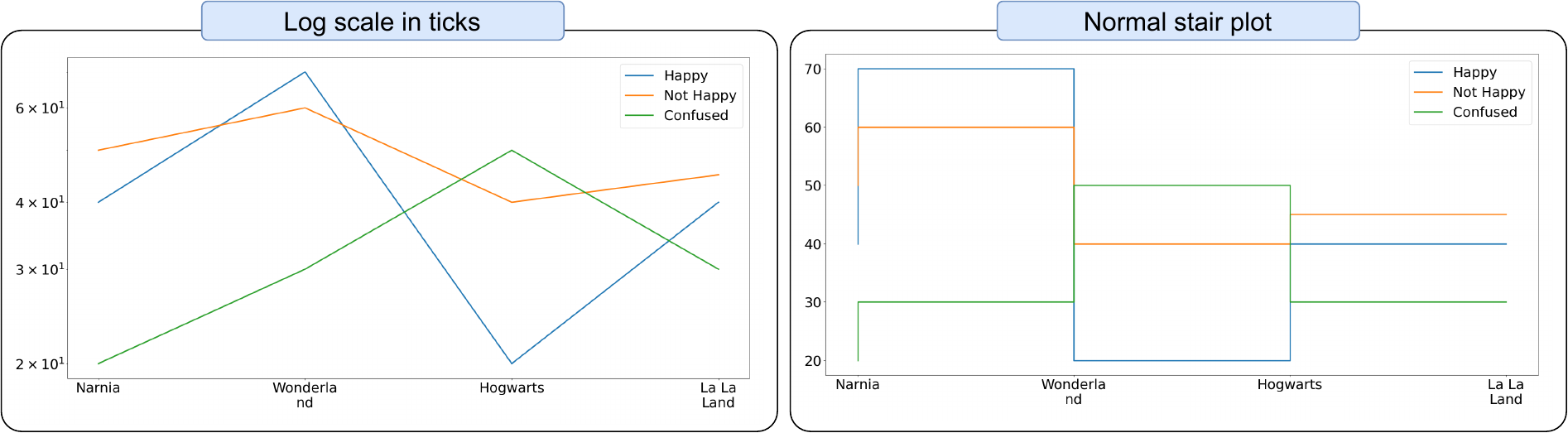}
    \label{fig:ls}
\end{figure*}

\begin{figure*}
    \centering
    \includegraphics[width = \linewidth]{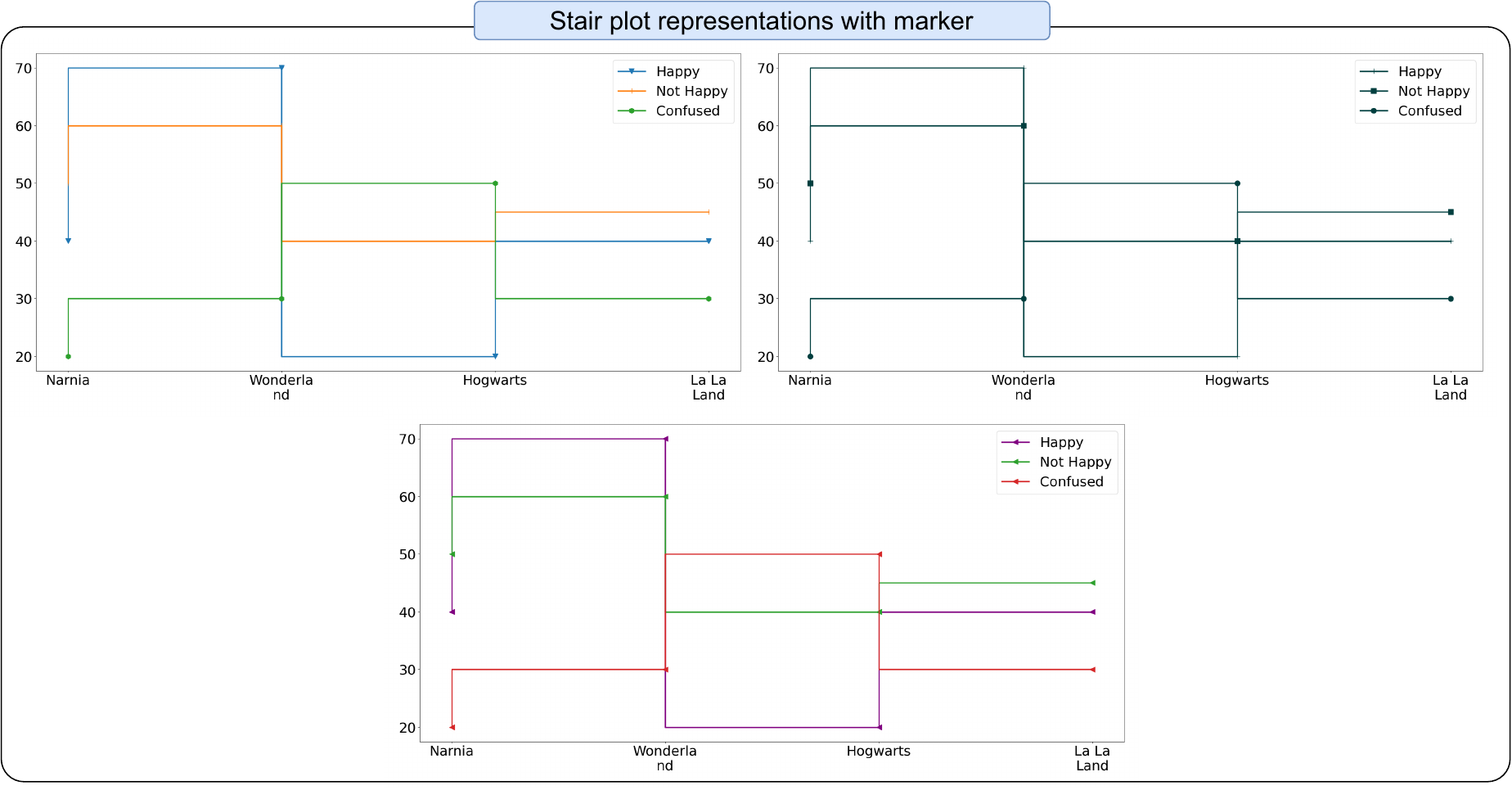}
    \label{fig:stair_marker}
\end{figure*}

\begin{figure*}
    \centering
    \includegraphics[width = \linewidth]{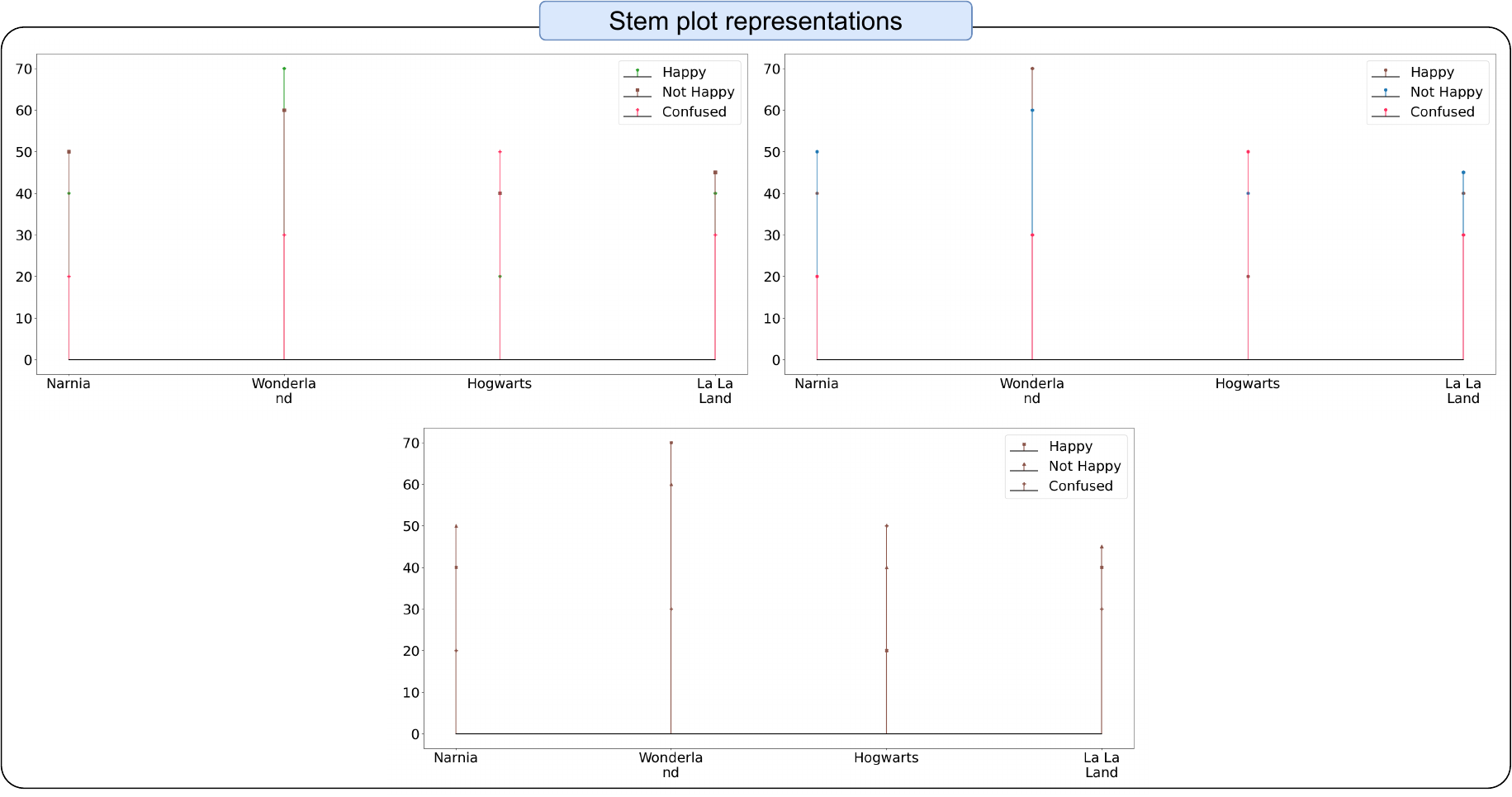}
    \label{fig:stair_pt}
\end{figure*}

\begin{figure*}
    \centering
    \includegraphics[width = \linewidth]{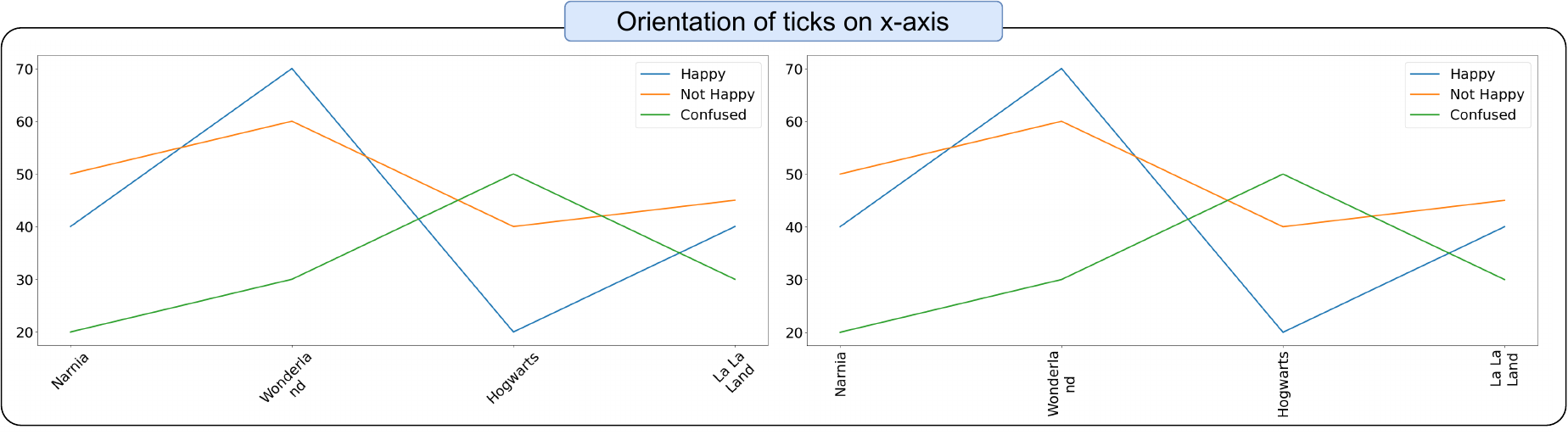}
    \label{fig:tick_orientatation}
\end{figure*}

\begin{figure*}
    \centering
    \includegraphics[width = \linewidth]{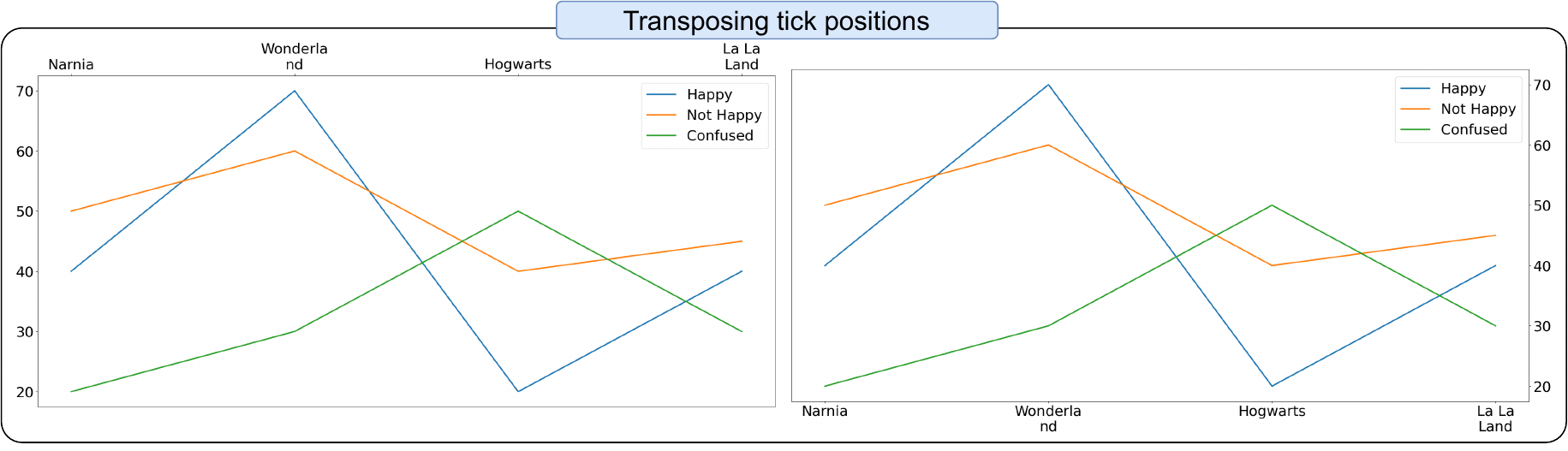}
    \label{fig:tick_pos}
\end{figure*}

\end{document}